\documentclass[lettersize,journal]{IEEEtran}
\usepackage{amsmath,amsfonts}
\usepackage{algorithmic}
\usepackage{algorithm}
\usepackage{array}
\usepackage[caption=false,font=normalsize,labelfont=sf,textfont=sf]{subfig}
\usepackage{textcomp}
\usepackage{stfloats}
\usepackage{url}
\usepackage{verbatim}
\usepackage{graphicx}
\usepackage{cite}
\usepackage{hyperref}

\usepackage{multirow}
\usepackage{booktabs}
\usepackage[accsupp]{axessibility}  

\begin{document}

\title{DreamCar: Leveraging Car-specific Prior for in-the-wild 3D Car Reconstruction}

\author{Xiaobiao~Du, Haiyang~Sun, Ming~Lu, Tianqing~Zhu, Xin~Yu
\IEEEcompsocitemizethanks{
This research is funded in part by ARC-Discovery grant (DP220100800 to XY) and ARC-DECRA grant (DE230100477 to XY). We thank all anonymous reviewers and editors for their constructive suggestions.

\IEEEcompsocthanksitem 
Xiaobiao Du is with the University of Technology Sydney, Australia (Email: xiaobiao.du@student.uts.edu.au)
\IEEEcompsocthanksitem
Haiyang Sun and Ming Lu are with Li Auto Inc. and Intel Inc., China, respectively. (Email: sunhaiyang@lixiang.com, ming1.lu@intel.com)
\IEEEcompsocthanksitem 
Tianqing Zhu is currently a professor at city university of Macau, China. (Email: tqzhu@cityu.edu.mo)
\IEEEcompsocthanksitem
Xin Yu is with The University of Queensland.
(Email: xin.yu@uq.edu.au)
\IEEEcompsocthanksitem Corresponding Author: Xin~Yu
}
}

\markboth{Journal of \LaTeX\ Class Files,~Vol.~14, No.~8, August~2021}%
{Shell \MakeLowercase{\textit{et al.}}: A Sample Article Using IEEEtran.cls for IEEE Journals}

\IEEEpubid{0000--0000/00\$00.00~\copyright~2021 IEEE}

\maketitle

\begin{abstract}
    Self-driving industries usually employ professional artists to build exquisite 3D cars. However, it is expensive to craft large-scale digital assets. Since there are already numerous datasets available that contain a vast number of images of cars, we focus on reconstructing high-quality 3D car models from these datasets. However, these datasets only contain one side of cars in the forward-moving scene. We try to use the existing generative models to provide more supervision information, but they struggle to generalize well in cars since they are trained on synthetic datasets not car-specific. In addition, The reconstructed 3D car texture misaligns due to a large error in camera pose estimation when dealing with in-the-wild images. These restrictions make it challenging for previous methods to reconstruct complete 3D cars. To address these problems, we propose a novel method, named DreamCar, which can reconstruct high-quality 3D cars given a few images even a single image. To generalize the generative model, we collect a car dataset, named Car360, with over 5,600 vehicles. With this dataset, we make the generative model more robust to cars. We use this generative prior specific to the car to guide its reconstruction via Score Distillation Sampling. To further complement the supervision information, we utilize the geometric and appearance symmetry of cars. Finally, we propose a pose optimization method that rectifies poses to tackle texture misalignment. Extensive experiments demonstrate that our method significantly outperforms existing methods in reconstructing high-quality 3D cars. \href{https://xiaobiaodu.github.io/dreamcar-project/}{Our code is available.}
    
\end{abstract}


\begin{figure*}[!ht] 
    \centering
    \includegraphics[width=1\textwidth]{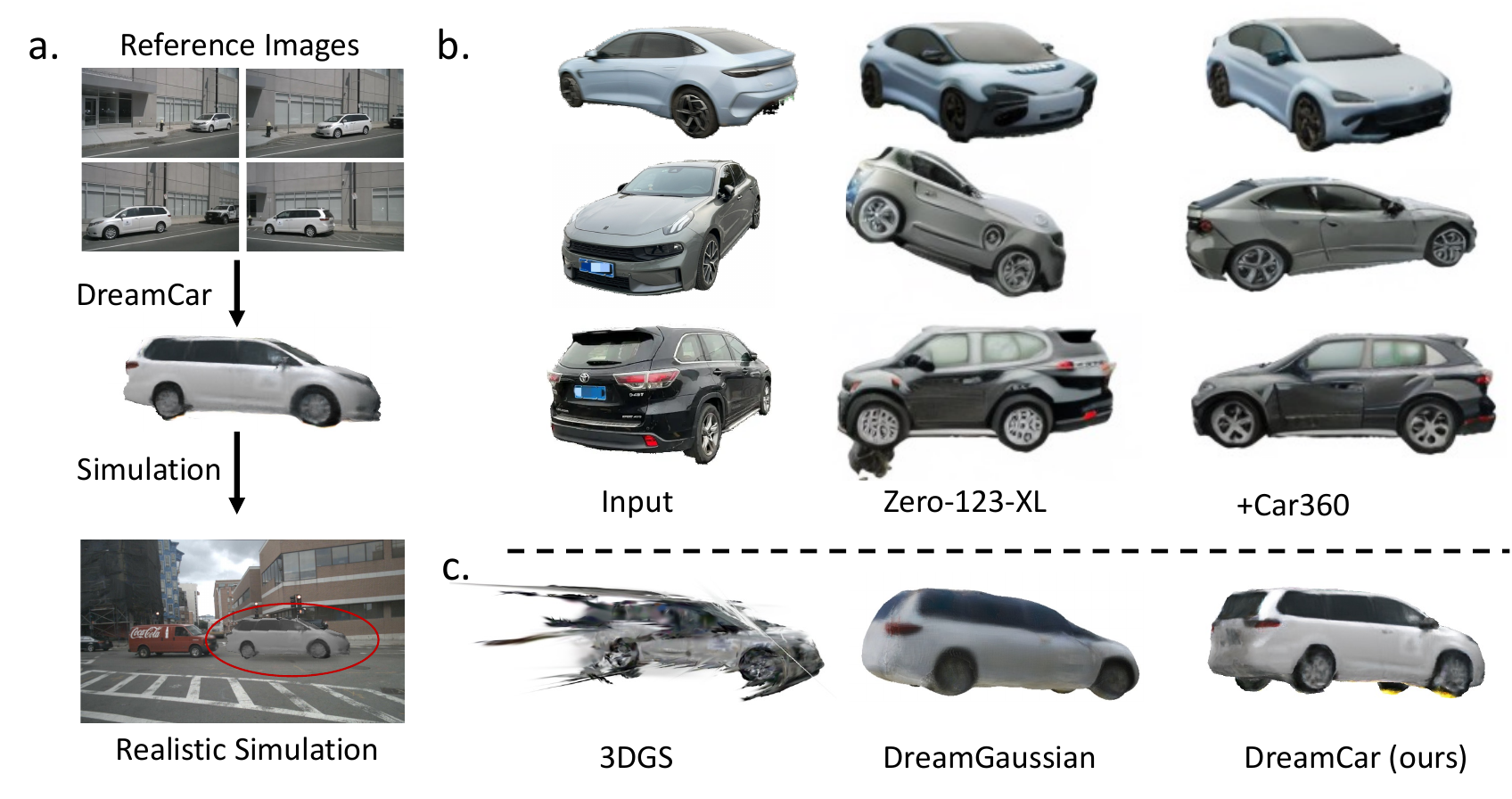} 
    \caption{\textbf{The illustration of the application scenario of our work and comparison with other methods.} \textbf{a.} Utilizing merely 4 reference images even less in the moving forward scene, we achieve the reconstruction of a complete 3D object, which is then simulated within a realistic scene. \textbf{b.} The novel view synthesis comparison provided by Zero-123-XL gradually trained on our Car360 dataset. \textbf{c.} The visual comparison of different 3D reconstruction methods. }
    \label{teaser}
    \vspace{-5mm}
\end{figure*}

\section{Introduction}

\IEEEPARstart{S}{elf-driving} vehicles are trained to safely interact with diverse environments and other vehicles. They must deal with infrequent dangerous situations, but collecting such dangerous data is challenging in the real-world setting~\cite{tan2021scenegen}. Simulation offers a flexible solution for generating large-scale data safely. For the development of a reliable self-driving system, it is essential to have training data that encompasses as broad a spectrum of scenarios as possible. This necessitates a simulator equipped with a diverse and extensive collection of traffic assets, such as vehicles of varying sizes, shapes, and appearances, to ensure comprehensive coverage.
Nevertheless, current self-driving simulators are limited as their 3D assets are created manually. The manual design of 3D assets is a labor-intensive and costly process, which limits scalability. Therefore, we intend to reconstruct large-scale high-quality 3D vehicle assets from existing self-driving datasets \cite{nuscenes, waymo, kitti}.

However, the process of producing such high-quality 3D cars from real-world sensor data faces 3 substantial challenges. (1) Captured Images that contain certain vehicles in self-driving datasets are often limited in number (ranging from one to five views). Particularly, in these images, only one side of the cars can be observed since they are captured in the moving forward scene. This scenario highlights a significant limitation: objects of interest are documented from restricted perspectives, with certain parts remaining unobserved. (2) Current large-scale 3D-aware diffusion models \cite{zero123, liu2023syncdreamer,stable-zero123} trained on large-scale synthetic datasets, not car-specific, generalize poorly in cars, especially real-world cars, which hinders them from providing useful prior. (3) Since the ego vehicle is in the moving forward scene and may encounter vibration, the estimated poses in these self-driving datasets contain a quite large error, leading to texture misalignment while reconstructing 3D cars.


In this work, we propose a novel method for 3D car reconstruction in the moving forward scene, named DreamCar, which utilizes existing real sensor data collected by self-driving vehicles in the moving forward scenario to reconstruct a high-quality 3D asset library for realistic sensor simulation.  
Considering the challenging moving forward situation that only provides one to five supervision images, our method introduces more supervision information. In light of the mirror symmetry of the nature of cars, the image flip and pose symmetry techniques are leveraged in our method to generate mirror counterparts, increasing the amount of supervision reference images to two times.  

To improve the generalization of the 3D-aware diffusion model to cars, we collect a high-quality car dataset, named Car360. As we can see in  Figure \ref{teaser} \textbf{b}, given input views, we use the current 3D-aware diffusion model, Zero-123-XL \cite{zero123} to synthesize novel views. Initially, this model struggles to synthesize realistic car novel views. However, after training with our collected Car360 dataset, which contains over 5,600 synthetic cars with photorealistic textures, the performance of this model on car images significantly improved. This improvement emphasizes the importance of our datasets and underscores existing 3D-aware models that generalize poorly in cars.

To tackle the texture misalignment problem raised by camera pose error, we propose a pose optimization method. This method can be integrated into our method without explicit supervision information. In particular, we design a Multilayer Perceptron (MLP), called PoseMLP, which takes the original pose and time-aware information as input to predict the offset for the correction to the original pose in the moving forward scenario.

As illustrated in Figure \ref{teaser} \textbf{a}, we show the effect of our proposed method, DreamCar, within its application context. Despite the challenge posed by training with only four low-resolution and noisy reference images, with our proposed techniques, DreamCar is capable of accurately reconstructing a complete 3D object using 4 reference images, with both precise geometry and intricate texture. This method proves valuable for reconstructing large-scale 3D vehicle objects from existing autonomous driving datasets and enables realistic simulation across a variety of scenes.
With our proposed techniques and dataset, Figure \ref{teaser} \textbf{c} demonstrates that our method can reconstruct more realistic 3D cars and our collected dataset bridges the gap of our reconstructed cars to cars. To sum up, our contributions can be summarized as follows:




$\bullet$ We propose DreamCar, a novel 3D car reconstruction method, tailored for the moving forward scene. Our method integrates mirror symmetry, generative prior, and pose optimization techniques, which enable our method to reconstruct an intact 3D object in the self-driving scene even if it is only given a single reference image.  

$\bullet$ To improve the generative model generalizing well in cars, we collect a Car dataset, termed Car360, with over 5,000 vehicles together, each adorned with photorealistic textures.

$\bullet$ To demonstrate our method, we evaluate our method in the large-scale self-driving dataset. We also evaluate it in our Car360 dataset.  Extensive experiments demonstrate our method outperforms existing methods and is effective in reconstructing large-scale complete and high-quality 3D car objects.

\section{Related Work}

\noindent\textbf{3D Reconstruction.}
The recent advancements  \cite{zhang2020nerf++, tancik2022block, xian2021space, boss2021nerd, wei2021nerfingmvs, reiser2023merf, yuan2022nerf, garbin2021fastnerf, park2021nerfies, chen2023mobilenerf, lin2021barf, tancik2023nerfstudio} in 3D reconstruction and novel view synthesis significantly propel progress in the field. A particularly noteworthy approach is the Neural Radiance Fields (NeRF) \cite{mildenhall2021nerf} which represents the entire scene as a radiance field parameterized by an MLP. This method employs the volume rendering \cite{volume_rendering} to render the scene. However, the applicability of NeRF is constrained to specific contexts \cite{mip-nerf360}, such as synthetic environments, and it is limited to scenes within its bounds, struggling with out-of-bound scenarios. Several methods are proposed, including Mip-NeRF \cite{mip-nerf}, Mip-NeRF 360 \cite{mip-nerf360}, TensoRF \cite{TensoRF}, and Instant-NGP \cite{instant-ngp}, all aimed at extending the utility of NeRF to more general, in-the-wild scenes.  The introduction of 3DGS \cite{kerbl2023gaussiansplatting} marks a further enhancement of the performance in 3D reconstruction, both in terms of quality and speed. Nonetheless, these methods encounter challenges in self-driving contexts, especially when tasked with reconstructing specific objects from a limited number of supervision images.

\noindent\textbf{Diffusion Models for 3D Generation.}
Recent advancements in 2D diffusion models \cite{stablediffusion, dhariwal2021diffusion}, as highlighted by notable works such as Stable Diffusion, open new avenues for the generation of 3D objects, marking a significant leap forward in the field. DreamFusion \cite{dreamfusion} and SJC \cite{sjc} pave the way by leveraging the capabilities of 2D text-to-image generation models to facilitate the creation of 3D shapes. This innovative approach inspires a wave of subsequent research, including text-to-3D methods
\cite{chen2023text,  seo2023ditto, yu2023points, seo2023let, tsalicoglou2023textmesh, armandpour2023re, chen2023it3d, xu2022neurallift,  cheng2023progressive3d}
and single-image-to-3D methods \cite{ qian2023magic123, zero123, yu2023hifi,shen2023anything, melas2023realfusion, tang2023make,  ling2023align, ma2023x}. 
To deal with the multi-head problem, there are several methods \cite{shi2023mvdream, li2023sweetdreamer, huang2023dreamcontrol, long2023wonder3d, wang2023prolificdreamer, zhang2023repaint123, szymanowicz2023viewset} that are proposed to exploit multi-view information. To tackle the generated results with unrealistic texture, more powerful 3D generation methods
\cite{liu2023one2345++,  shi2023zero123plus, raj2023dreambooth3d, chen2023fantasia3d,liu2024one} and improved diffusion strategies \cite{lin2023magic3d, liu2023unidream, zhao2023efficientdreamer, zhu2023hifa, huang2023dreamtime, ma2023x, wu2023hd,jiang2023efficient, chen2023et3d} are proposed.   To speed up the generation, some methods \cite{ yi2023gaussiandreamer, ling2023align, liu2023humangaussian, chung2023luciddreamer, xu2024agg, szymanowicz2023splatter} exploit Gaussian Splatting \cite{kerbl2023gaussiansplatting} as the main 3D model. Specifically, Dreamgaussian \cite{tang2023dreamgaussian}, as the representative Gaussian Splatting generating method, generates a 3D object in 40 seconds. 
On the contrary, for better generation, DreamCraft3D \cite{sun2023dreamcraft3d} adopts multi-stage training to achieve state-of-the-art generation results.
Nevertheless, the application of these methods to realistic scenarios presents challenges, primarily because they heavily depend on generative models. This reliance often results in generated outcomes that lack plausibility when compared to realistic references.

\section{Proposed Car360 Datasets}

\begin{figure*}[t] 
    \centering
    \includegraphics[width=1\textwidth]{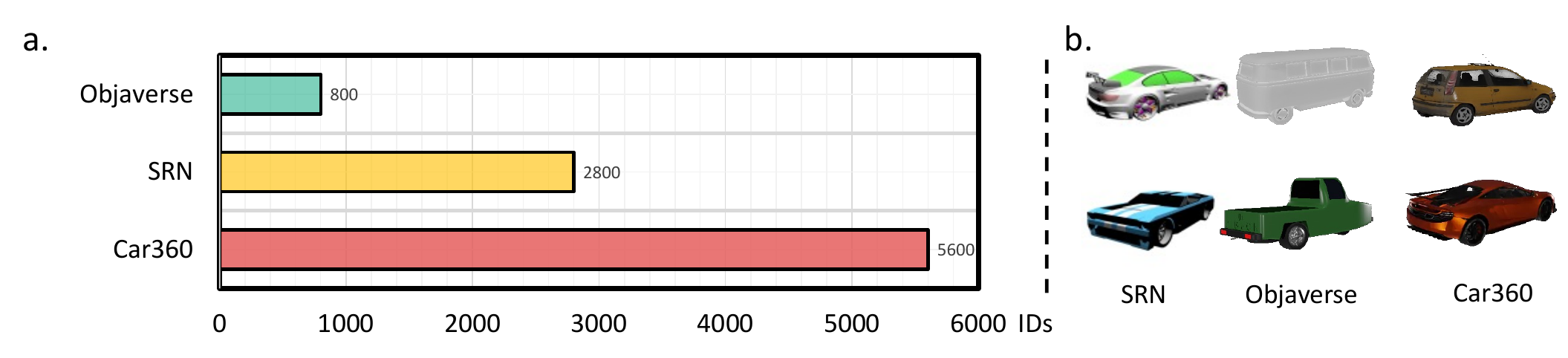} 
    \caption{\textbf{The illustration of different car datasets.} \textbf{a.} The comparison of the total number of vehicles across different datasets, highlights that our Car360 collection has the highest count, with 5,600 vehicles.  \textbf{b.} The visual comparison of different car datasets. \textbf{c.} The illustration of the distribution of various vehicle categories, lighting conditions, and the number of captured views in our collected Car360 dataset. }
    \label{dataset}
    \vspace{-5mm}
\end{figure*}

This work aims to reconstruct a complete 3D model from a limited number of images, typically ranging from one to five.  However, relying solely on this supervision information is insufficient. Therefore, we integrate a generative prior from the recent large-scale 3D-aware diffusion model, Zero-123-XL \cite{zero123} in our method. We found this model fails to generalize well in the realistic car subject as shown in Figure \ref{teaser} \textbf{b}, attributed to its training on large-scale synthetic datasets, like Objaverse \cite{deitke2023objaverse, deitke2024objaverse}, not car-specific.  In this work, we collect a car dataset, named Car360, which contains 5,600 synthetic cars to enable our model to focus on cars and boost our model robust to realistic cars. Figure \ref{dataset} \textbf{b} shows the reality comparison among different datasets.

As shown in Figure \ref{dataset} \textbf{a} and \textbf{b}, we list the number of vehicles from the existing car dataset with 360-degree views and show some samples from them. We find that a large number of samples from SRN \cite{chang2015shapenet} and Objaverse \cite{deitke2023objaverse, deitke2024objaverse} are not realistic. 
Therefore, we collect a synthetic car dataset from 
Sketchfab \footnote{Sketchfab: \href{https://sketchfab.com}{https://sketchfab.com}}, consisting of 2,000 3D vehicles, named Car360. Binding them with SRN and Objeaverse forms our Car360 dataset. This dataset contains more realistic car 3D models than other synthetic datasets. The generative model would be trained on this dataset to improve its robustness to cars.

\section{Method}

\begin{figure*}[t] 
    \centering
    \includegraphics[width=1\textwidth]{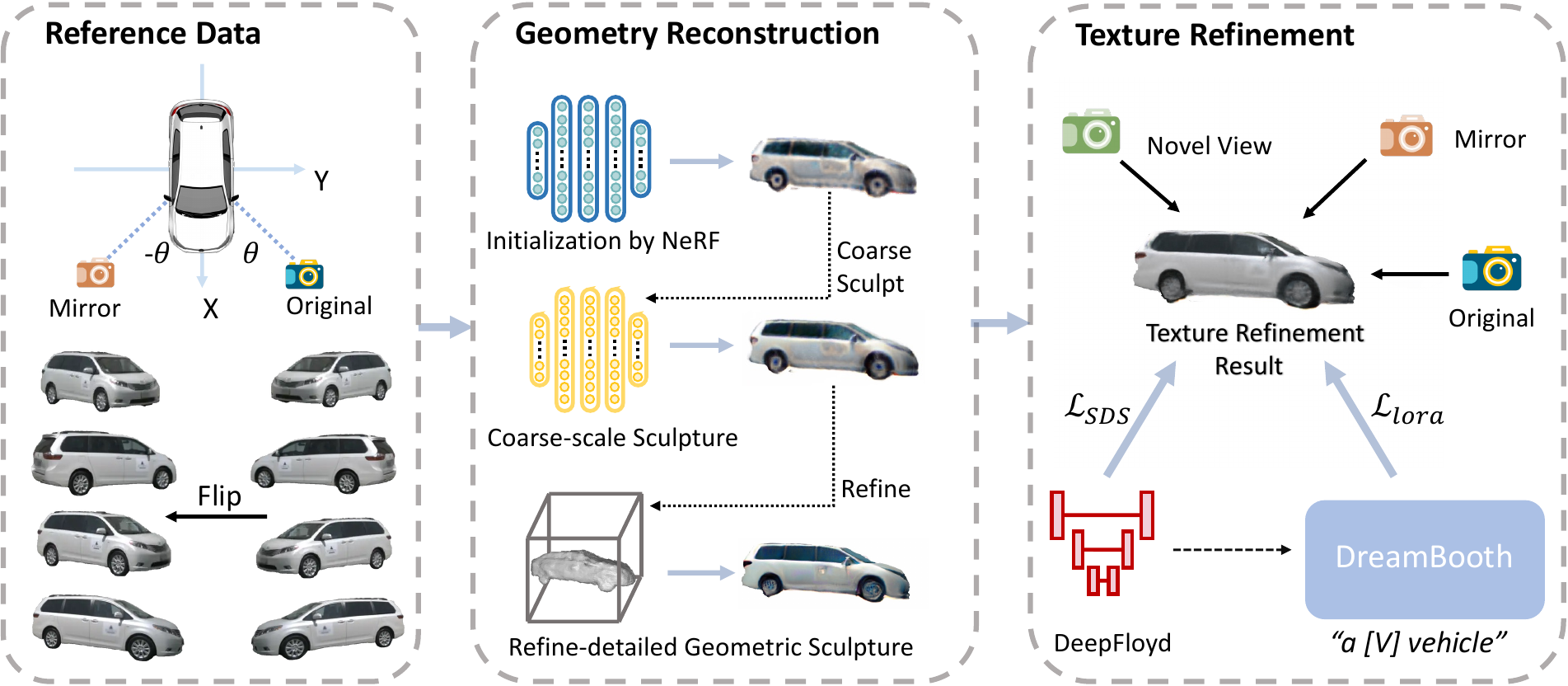} 
    \caption{
\textbf{The illustration of our DreamCar. }Our method can be divided into geometry reconstruction and texture refinement stages. We input reference views (original reference views and their mirror counterparts) with generative prior guiding our 3D model to reconstruct 3D cars in all stages. In the geometry reconstruction stage, our method progressively sculpts fine geometry with coarse texture. In the texture refinement stage, we focus on the refinement of the appearance of the car with the DreamBooth technique.}
    \label{method}
    \vspace{-5mm}
\end{figure*}

\subsection{Preliminaries}

\subsubsection{Score Distillation Sampling.}
Score Distillation Sampling (SDS), as described in DreamFusion \cite{poole2022dreamfusion}, is a representative method to optimize 3D representation via a pre-trained 2D diffusion model $\phi$, like stable-diffusion \cite{stablediffusion}. 
In DreamFusion, MipNeRF \cite{barron2021mip} is adopted as the 3D representation with parameters $\theta$ subject to optimization. With $g$ denoting the rendering function, the image produced, $ I_r = g(\theta)$, emerges from this process. 
To make the rendered image $I_r$ look like the sample generated from the diffusion model $\phi$, SDS is proposed to leverage the generated prior from the diffusion model. SDS exploits the diffusion model to predict the sampled noise $\hat\epsilon_\phi$ given the noisy image $z^r_t$, text embedding $y$, and time step $t$, as an estimated score $\hat\epsilon_\phi(z^r_t; y, t)$.
The method assesses the deviation between the rendered image $I_r$ added with Gaussian noise $\epsilon$ to the and another noise $\hat\epsilon_\phi$ predicted by the diffusion model. This measurement guides the adjustment of the parameters $\theta$. The process of calculating the gradient for this adjustment is described as follows:

\begin{equation}
    \nabla_{\theta} \mathcal{L}_{SDS}(\phi, g(\theta)) \triangleq \mathbb{E}_{t, \epsilon}[w(t)\left(\hat\epsilon_\phi(z^r_t; y, t)  - \epsilon\right) {\partial I_r \over \partial \theta}],
\end{equation}
where $w(t)$ is a weighting function. However, this method using the prompt with a certain view instruction to drive the diffusion model to generate a fixed view is rigid and only generates coarse views. To address this concern, Zero-123 \cite{zero123} is proposed to exploit a conditioning image $I_{cond}$ with a relative 3 Degrees of Freedom (DoF) pose $\Delta p$ to generate a novel view. The integration of Zero-123 into SDS can be formulated as follows:

\begin{equation}
    \nabla_{\theta} \mathcal{L}_{3D-SDS}(\phi, g(\theta)) \triangleq \mathbb{E}_{t, \epsilon}[w(t)\left(\hat\epsilon_\phi(z^r_t; I_{cond}, \Delta p, t)  - \epsilon\right) {\partial I_r \over \partial \theta}].
\end{equation}

\subsection{Camera Pose Processing}
SfM \cite{sfm2016} is a common method to reconstruct point clouds and estimate the camera pose according to a series of images in the real world. However, cameras on the ego vehicle only capture a few images of a certain object, making SfM fail to estimate the camera pose. Thus, we rely on the large-scale self-driving Object Detection dataset, Nuscenes \cite{nuscenes} with clean annotations to extract the pose of captured vehicles.
Given the camera poses $p_{cam}$ and ego poses $p_{ego}$ from the camera in the dataset like Nuscenes \cite{nuscenes}, these poses can all be described by Lie group $SE(3)$ \cite{mishra2013lie}. The Lie group can be formulated as follows:

\begin{equation}
SE(3) = \left\{\begin{bmatrix} \mathbf{R} & \mathbf{T} \\
\mathbf{0}^T & 1 \end{bmatrix} \in \mathbb{R}^{4 \times 4}|\mathbf{R} \in SO(3),\mathbf{T} \in \mathbb{R}^{3}\right\},
\end{equation}
where  $\mathbf{R}$ and $\mathbf{T}$ denote the rotation matrix and translation. With these two poses, we can easily get the camera-to-world pose $p_{c2w} = p_{ego} p_{cam}. $
Since we target to reconstruct an object, not the whole scene, so we need to transfer the current reference system to ``object-centric''. We extract the pose of a certain object $p_{obj}$ by using the bounding box annotation. The central point of the 3D bounding box and its orientation can represent the translation and rotation matrix as the pose of an object, as $p_{obj}$. Therefore, we transfer the camera-to-world pose to ``object-centric'' $p_{c2obj} = p_{obj}^{-1} p_{c2w}$. The inverse of the pose can be calculated by:

\begin{equation}
SE(3)^{-1} = \left\{\begin{bmatrix} \mathbf{R}^T & -\mathbf{R}^T\mathbf{T} \\
\mathbf{0}^T & 1 \end{bmatrix} \in \mathbb{R}^{4 \times 4}|\mathbf{R} \in SO(3),\mathbf{T} \in \mathbb{R}^{3}\right\}.
\end{equation}

Since we aim at reconstructing a 3D object in large-scale self-driving datasets, the number of images of a certain object is about one to five, making existing methods fail to reconstruct a high-quality 3D object. Thus, we reconstruct a 3D object by means of the generated prior from SDS. SDS replies on the large-scale diffusion models \cite{stablediffusion, zero123} to provide generated information. However, these generated models all assume the object is in the center of an image. Therefore, we recenter the input images and adjust the camera pose to look at the original point. The final pose $p$ would be obtained by adjusting the rotation matrix of pose $p_{c2obj}$ to look at the coordinate original point.


\subsection{Pre-processing and Attainment of Mirror Symmetry}
Given a few images $I = \left\{I_1, I_2, \cdots I_n \right\}$ each capturing a certain vehicle from an autonomous driving dataset with the number of images $|I|$ ranging from one to five, our objective is to reconstruct a 3D object characterized by precise geometry and clean texture. Remarkably, our method is capable of reconstructing a high-quality 3D model from a few images even a single image, despite the challenging constraints posed by such few supervision data. Our initial step involves employing the Segment Anything Model (SAM) \cite{sam} to segment the vehicle from the background, focusing our reconstruction solely on the vehicle rather than the entire scene. To address the scarcity of supervision information, we expand this limited training data by exploiting mirror symmetry to enrich the information base for the reconstruction process.

There are plenty of objects that have symmetry in nature, and vehicles are no exception. 
Generally speaking, the left and right side of a car has the same appearance and keep symmetry. Considering this significant feature, we first expand the number of reference images by the mirror flip. Denoting the image flip function as $F_{flip}$, the mirror symmetry image $I_i^{\prime}$ can be obtained by simply flipping the image $I_i^{\prime} = F_{flip}(I_i)$. The leftmost part of Fig. \ref{method} demonstrates how we obtain the image and pose mirror. To obtain the mirror camera pose, taking the camera pose of the Colmap format \cite{sfm2016} for instance, the car orients along the x-axis, so we just need to negative the y component of the translation and mirror the rotation matrix.  With the obtained mirror images and poses, the reference images can be expanded to two times $I = \left\{I_1, I_2, \cdots I_n, I_1^{\prime}, I_2^{\prime}, \cdots I_n^{\prime} \right\}$.



\subsection{Geometry Reconstruction}
As shown in Fig. \ref{method}, our method can be divided into the geometry reconstruction and texture refinement stages.    In the geometry reconstruction stage, we focus on sculpting a 3D object with intact and precise 3D geometry and coarse texture such that it matches reference images $I$ at reference views and maintains plausibility at various views. 

To reach this target, we choose to progressively sculpt fine structural geometry by three different 3D models.
NeRF is first adopted to reconstruct a coarse geometry and shape. Then, we use the result of NeRF as initialization for the Neus \cite{wang2021neus}. Once the coarse-scale sculpture by Neus is completed, we use DMTET \cite{shen2021deep} initialized by Neus to reconstruct a fine geometry. To leverage most of the absolute supervision information provided by the reference image, we penalize for the foreground discrepancy observed between the rendered image and the reference and their mask through

\begin{equation}
     \mathcal{L}_{\textnormal{rgb}} = \left\lVert {m_i} \odot \left(I_i - g(\theta; p_i)\right) \right\rVert_2, \quad
      \mathcal{L}_\textnormal{mask} = \left\Vert {m_i} - g_m(\theta; p_i)\right\Vert_2,
\end{equation}
at the $i$-th reference pose $p_i$. We only compute the loss within the foreground region through the reference mask ${m_i}$. At the same time, we also compute the mask loss to encourage the outline convergence, where $g_m$ renders the silhouette. In addition, drawing inspiration from \cite{deng2023nerdi}, we compute both depth and normal losses to thoroughly utilize the geometric prior derived from the reference image, thereby guaranteeing geometry matching with the reference.  The depth and normal loss can be computed as:
 
\begin{equation}
    \mathcal{L}_\textnormal{depth}= -\frac{\text{cov}({d_i},  g_d(\theta; p_i))}{\sigma({d_i}) \sigma(g_d(\theta; p_i))},
    \quad \mathcal{L}_\textnormal{normal}= -\frac{{n_i} \cdot g_n(\theta; p_i)}{{\Vert {n_i} \Vert}_2 \cdot {\Vert g_n(\theta; p_i) \Vert}_2},
\end{equation}
where $\text{cov}(\cdot)$ and $\sigma(\cdot)$ denote the covariance and variance operators respectively. We obtain the reference depth $d_i$ and the normal $n_i$ by leveraging the off-the-shelf network \cite{eftekhar2021omnidata}. 
The rendered depth and normal are obtained by $g_d(\theta; p_i)$ and $g_n(\theta; p_i)$.
To the depth loss, we adopt the negative Pearson correlation $\mathcal{L}_\textnormal{depth}$ to address discrepancies in the depth scale. To normal loss, we simply compute its cosine similarity to ensure the fitting of the rendered normal and the reference one. 
However, this supervision information is not enough for the reconstruction of a complete 3D object. Thus, we employ the generative prior to enhance the reconstruction of distinct views given references.

\noindent\textbf{Generative Prior.} The recent 3D-aware generative model, Zero-123 \cite{zero123}, has been trained on a vast array of 3D objects, endowing it with an innate understanding of 3D poses up to three degrees of freedom (3DoF). In this work, we leverage our collected Car360 dataset to enhance the generalization of the model to real-world cars. This model is then employed to generate detailed information that aids in the reconstruction of a comprehensive 3D model.
Moreover, we incorporate a 2D generative model to augment the generative capabilities, given its training on a broader and more diverse dataset compared to those available to 3D-aware generative models. Consequently, we employ a mixed SDS loss that mixes the generative priors from both 2D and 3D perspectives. This mixed SDS loss is utilized to distill the gradient as follows:
\begin{equation}
\nabla_{\theta}\mathcal{L}_{mix}(\phi, g(\theta))=\nabla_{\theta}\mathcal{L}_\textnormal{SDS}(\phi, g(\theta)) + \nabla_{\theta}\mathcal{L}_\textnormal{3D-SDS}(\phi, g(\theta)).
\end{equation}
When computing $\mathcal{L}_\textnormal{SDS}$, the DeepFloyd IF base model \cite{deep-floyd} \cite{stablediffusion} is adopted as 2D prior to capture coarse geometry. Therefore, the total geometry reconstruction loss can be formulated as:

\begin{equation}
\begin{split}
         \mathcal{L}_{geo} &= \lambda_{rgb} \mathcal{L}_{rgb}
    + \lambda_{mask} \mathcal{L}_{mask}
    + \lambda_{depth} \mathcal{L}_{depth} \\
    &+ \lambda_{normal} \mathcal{L}_{normal}
    + \lambda_{mix} \mathcal{L}_{mix},
\end{split}
\end{equation}
where $\mathcal{L}_{rgb}, \mathcal{L}_{mask}, \mathcal{L}_{depth}, \mathcal{L}_{normal}$ are reference losses, and $\mathcal{L}_{mix}$ means the guidance loss introducing generative prior.  All $\lambda$ are weights for all losses.

\begin{table*}[!ht]
\caption{\textbf{The comparison of quantitative 3D reconstruction metrics in the test set from the Car360 dataset.} ``Standard'' means the evaluation in standard testing views. ``Mirror'' denotes the evaluation of the mirror version of the testing view to measure the completeness of reconstructed 3D cars. 
Infer. Speed (n/s) means the number of rendered images per second.
The best results are highlighted in bold.
}
\label{t1}
\centering
\begin{tabular}{c|l|ccccc}
\hline
\multicolumn{1}{l|}{Testing View} & Method                                                  & \multicolumn{1}{l}{MSE $\downarrow$} & PSNR $\uparrow$ & SSIM $\uparrow$ & LPIPS $\downarrow$ & FID $\downarrow$ \\ \hline
\multirow{4}{*}{Standard}                       
                                  & TensoRF \cite{TensoRF}                 & 0.1224                               & 9.23            & 0.3685          & 0.5932             & 0.91                                                         \\
                                  & Instant-NGP \cite{instant-ngp}         & 0.1127                               & 9.48            & 0.3837          & 0.5491             & 0.87                                                         \\
                                
                                  & 3DGS \cite{kerbl2023gaussiansplatting} & 0.0933                               & 10.06           & 0.4027          & 0.5022             & 0.67                                                         \\ \hline
                                  & \textbf{DreamCar (Ours)}                                & \textbf{0.0297}                      & \textbf{15.44}  & \textbf{0.6894} & \textbf{0.2281}    & \textbf{0.23}                                              \\ \hline
\multirow{4}{*}{Mirror}                          
                                  & TensoRF \cite{TensoRF}                 & 0.2367                               & 6.27            & 0.3092          & 0.6121             & 1.92                                                            \\
                                  & Instant-NGP \cite{instant-ngp}         & 0.1929                               & 7.21            & 0.3281          & 0.6075             & 1.59                                                           \\
                                  & 3DGS \cite{kerbl2023gaussiansplatting} & 0.1467                               & 7.83            & 0.3331          & 0.5904             & 1.31                                                       \\ \cline{2-7} 
                                  & \textbf{DreamCar (Ours)}                                & \textbf{0.0429}                      & \textbf{14.71}  & \textbf{0.6761} & \textbf{0.2591}    & \textbf{0.73}                                           \\ \hline
\end{tabular}
\end{table*}

\begin{table*}[!ht]
\caption{
\textbf{The comparison of point clouds and detection metrics in the test set from the Car360 dataset.} The best results are highlighted in bold.
}
\label{t2}
\centering
\begin{tabular}{l|c|c|c|c|c}
\toprule
Geometry    & 3DGS  \cite{kerbl2023gaussiansplatting}  & Zero-123 \cite{zero123} & DreamGaussian \cite{tang2023dreamgaussian} & DreamCraft3D \cite{sun2023dreamcraft3d} & \textbf{Dreamcar}         \\ \midrule \midrule
$L_2$-error $\downarrow$ & 0.24  & 0.19   & 0.15       & 0.14       & \textbf{0.13}  \\
Hit rate  $\uparrow$  & 66.38\% & 86.19\%  & 89.21\%       & 91.37\%      & \textbf{93.48\%} \\
Chamfer   $\downarrow$  & 9.66    & 2.14     & 0.87          & 0.41         & \textbf{0.24}    \\
Hausdorff  $\downarrow$   & 1.8418  & 1.28     & 1.45          & 1.39         & \textbf{0.98}    \\ 
map@0.5  $\uparrow$   & 0.04 & 0.23     & 0.31         & 0.67         & \textbf{0.83}    \\ \bottomrule
\end{tabular}
\end{table*}

\subsection{Texture Refinement}
The initial stage of geometry reconstruction yields precise geometry, yet the resulting textures are overly smooth and lack sharpness. This issue stems from the reliance on our 2D generative prior model, which generates images at a low resolution and exhibits considerable inconsistency when producing the specified subject. In this stage, we would focus on the refinement of texture.

To enhance the realism of the textures, we integrate the Stable Diffusion model \cite{stablediffusion} at this stage to provide gradients of higher resolution. Additionally, during this learning phase, we continue to utilize the aforementioned reference loss, given its value as an effective source of supervision information. To improve the stability of the generative model, we utilize the recently proposed DreamBooth \cite{ruiz2023dreambooth} and Lora \cite{wang2023prolificdreamer, hu2021lora} techniques.  Specifically, we set the text prompts containing the class of vehicles and a unique identifier (e.g., “a [V] vehicle” in Fig. \ref{method}) to enable the generation of the diffusion model stable to an identity. To the problem of blurry and over-statute texture, we introduce Lora, a large-scale model fine-tuning technique, which significantly enhances texture realism during the gradient distillation phase. Consequently, at this stage, we incorporate two additional loss functions, $L_{dreambooth}$ and $L_{lora}$, to achieve a more realistic texture generation.

\noindent\textbf{Pose Optimization.} 
In the moving forward scene, the estimated pose to a certain captured object exists a quite large error since it is in a dynamic environment and may encounter vibration. This problem would lead to texture misalignment in the process of the reconstruction of a fine 3D object. To tackle this problem, we propose a pose optimization method, which accepts the time frame and the original pose as input to predict the offset for the correction to the original pose. Specifically, we design an MLP, called PoseMLP $N_{posemlp}$, with 3 layers of 256 hidden neural units. To the rotation matrix, we do not simply input the whole one but rather quantify the rotation of each axis as input. Therefore, the total input for this network is 6D vectors and the time frame.
Our pose optimization can be formulated as:
\begin{equation}
    \hat{p_i} = N_{posemlp}(i, p_i) + p_i,
\end{equation}
where $p_i$ denotes the $i$-th pose. With this time-related information as input, this network is endowed with time-aware ability in the moving forward scene. Note that our proposed PoseMLP can be integrated into our method and does not need explicit supervision information. Therefore, the texture refinement loss can be represented as 
\begin{equation}
\begin{split}
       \mathcal{L}_{tex} &= \lambda_{rgb} \mathcal{L}_{rgb}
    + \lambda_{mask} \mathcal{L}_{mask}
    + \lambda_{depth} \mathcal{L}_{depth} 
    + \lambda_{normal} \mathcal{L}_{normal} \\
   & + \lambda_{3d} \mathcal{L}_{3D-SDS}  
    + \lambda_{booth} \mathcal{L}_{dreambooth}  
    + \lambda_{lora} \mathcal{L}_{lora},  
\end{split}
\end{equation}
where $\lambda_{rgb}$ would be the largest one since it provides absolution appearance supervision for better texture refinement.

\begin{figure*}[ht] 
    \centering
    \includegraphics[width=1\textwidth]{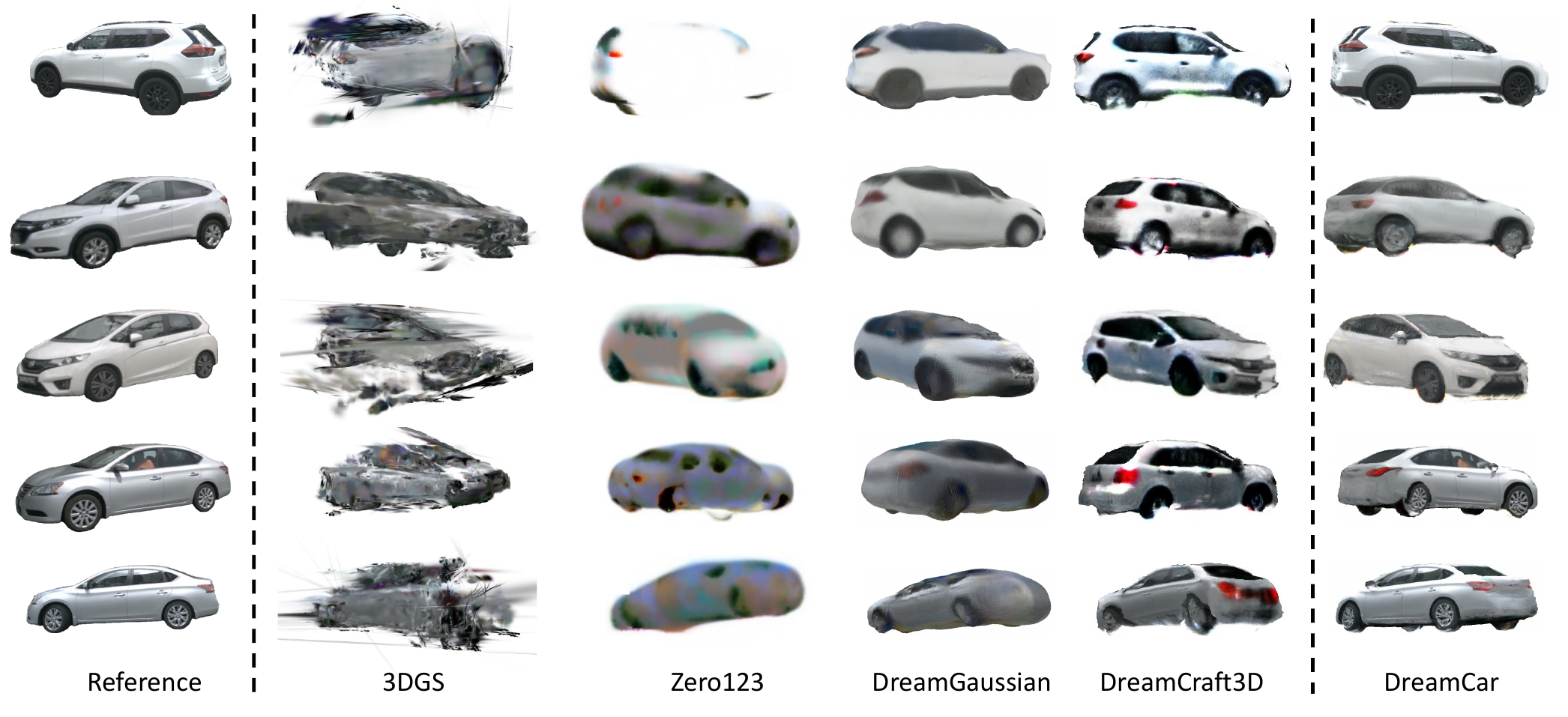} 
    \caption{\textbf{Qualitative evaluation of 3D reconstruction on the Nuscenes Dataset \cite{nuscenes}. }The renderings are provided from various viewpoints distinct from the reference image, illustrating the completeness of the reconstructed 3D models. }
    \label{vis_nus}
\end{figure*}

\section{Experiments}

\begin{table*}[t]
\caption{
\textbf{The comparison of quantitative 3D reconstruction metrics in 100 vehicles from the Nuscenes dataset.} ``Standard'' means we randomly choose one image of each vehicle as the test image. ``Mirror'' denotes the mirror version of the test image, employed to verify the completeness of the 3D model. The best results are highlighted in bold.
}
\centering

\label{t3}

\begin{tabular}{c|l|lcccc}

\toprule 
\multicolumn{1}{l|}{Testing View}     & Method      & MSE    $\downarrow$         & PSNR $\uparrow$          & SSIM      $\uparrow$      & LPIPS    $\downarrow$       & FID   $\downarrow$          \\ \midrule \midrule
\multirow{5}{*}{Standard} & NeRF  \cite{mildenhall2021nerf}   &    0.1341             &    8.74            &     0.3557            &     0.6019            &      1.31           \\
 & TensoRF  \cite{TensoRF}   &    0.1224             &    9.23            &     0.3685            &     0.5932            &      0.91           \\
                          & Instant-NGP \cite{instant-ngp}                &    0.1127            &   9.48              &    0.3837             &   0.5491      & 0.87        \\
                           & Zip-NeRF \cite{barron2023zipnerf}                &    0.1093            &   9.72              &    0.3861             &   0.5134     & 0.52       \\
                          & 3DGS   \cite{kerbl2023gaussiansplatting}     & 0.0933          & 10.06          & 0.4027          & 0.5022          & 0.67          \\ \cline{2-7} 
                          & \textbf{DreamCar (Ours) }   & \textbf{0.0297} & \textbf{15.44} & \textbf{0.6894} & \textbf{0.2281} & \textbf{0.23} \\ \midrule

\multirow{5}{*}{Mirror}   & NeRf  \cite{mildenhall2021nerf}   &  0.2481               &     5.13           &   0.2714              &     0.6934            &     2.36            \\                          
   & TensoRF  \cite{TensoRF}   &  0.2367               &     6.27           &   0.3092              &     0.6121            &     1.92            \\
                          & Instant-NGP  \cite{instant-ngp} &  0.1929               &   7.21             &  0.3281               &   0.6075              &  1.59               \\
                          & Zip-NeRF  \cite{barron2023zipnerf} &  0.1541               &   7.42             &  0.3357               &   0.6012              &  1.45               \\
                          & 3DGS \cite{kerbl2023gaussiansplatting}         & 0.1467          & 7.83           & 0.3331          & 0.5904          & 1.31         \\ \cline{2-7} 
                          & \textbf{DreamCar (Ours) }   & \textbf{0.0429}         & \textbf{14.71} & \textbf{0.6761} & \textbf{0.2591} & \textbf{0.73} \\ \bottomrule
\end{tabular}
\end{table*}
\begin{figure*}[t] 
    \centering
    \includegraphics[width=0.8\textwidth]{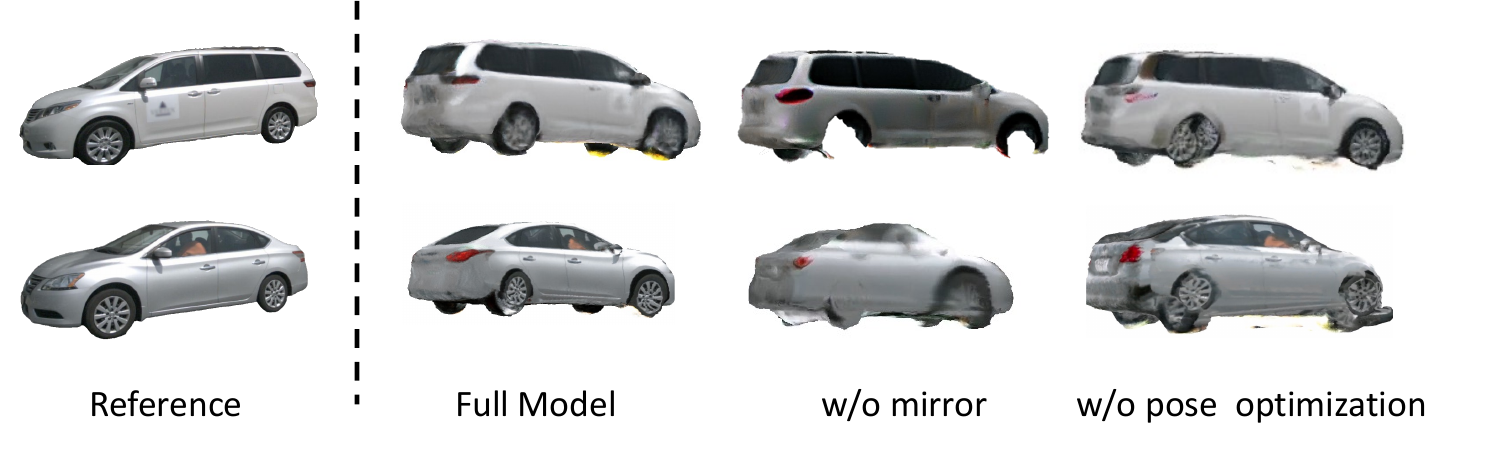} 
    \caption{\textbf{The ablation study of our proposed method.} Given the leftmost references, we ablate the mirror and pose optimization techniques to demonstrate our method.}
    \label{ablation}
\end{figure*}

\begin{figure*}[t] 
    \centering
    \includegraphics[width=0.8\textwidth]{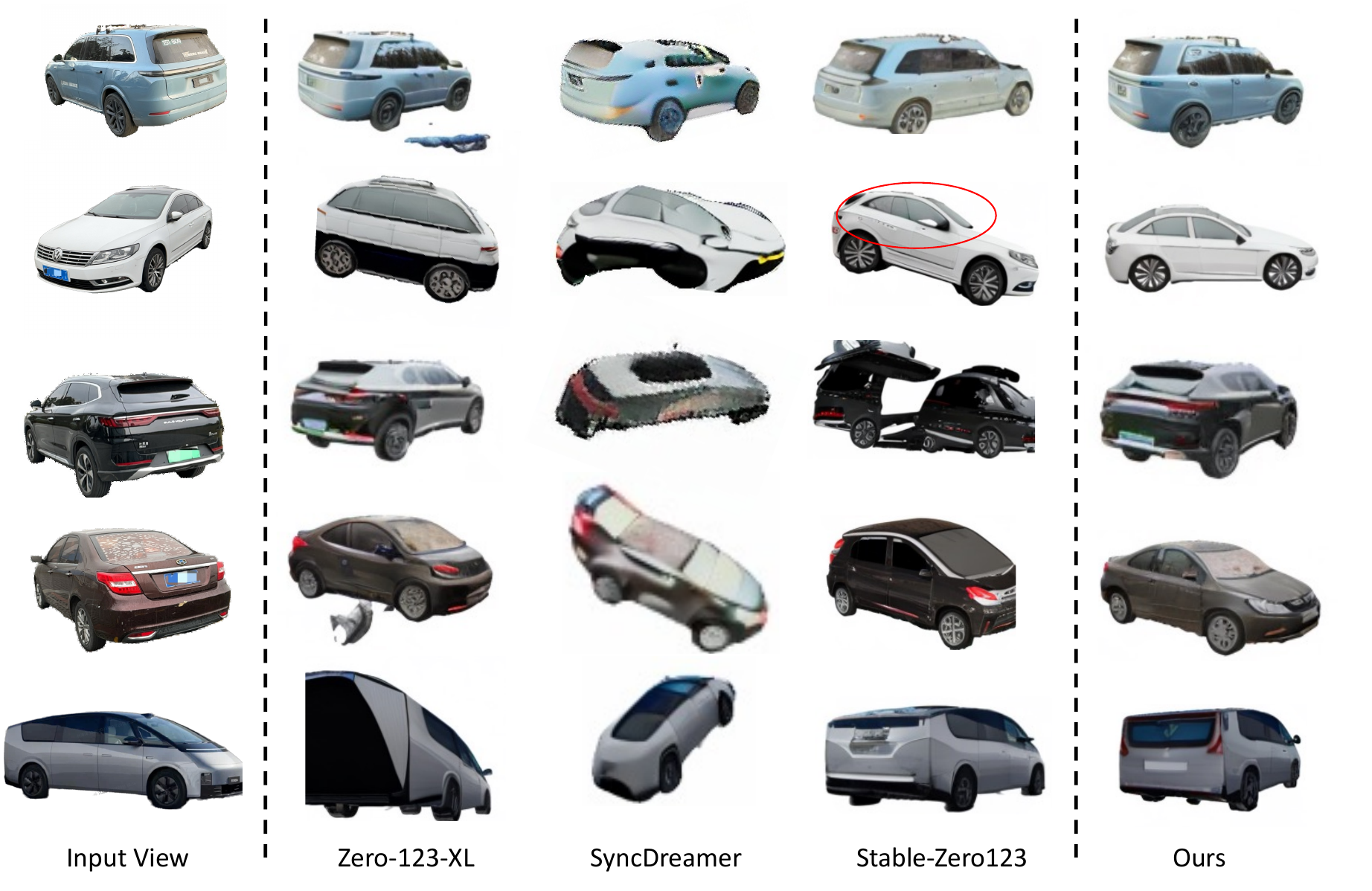} 
    \caption{\textbf{The comparison of novel view synthesis across 3D-aware diffusion models. }The red circle highlights the unreasonable generation of Stable-Zero123 \cite{stable-zero123}.}
    \label{vis_nvs}
\end{figure*}

\begin{figure*}[ht] 
    \centering
    \includegraphics[width=0.8\textwidth]{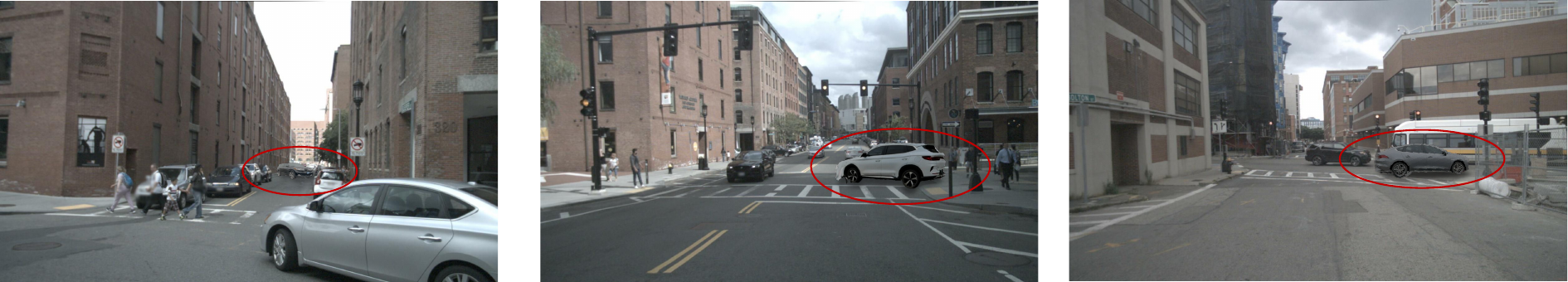} 
    \caption{\textbf{Car simulation in the original Nuscenes scenes.} The red circle indicates the simulated car.}
    \label{vis_simu}
\end{figure*}

\subsection{Implementations}

During the geometry reconstruction stage, we opt for a low resolution of 128 for both NeRF and Neus, aiming to expedite training time while achieving a coarse geometric structure. In the subsequent stages of geometry and texture refinement, we increase the resolution to 1024. This adjustment is designed to capture more intricate details in both geometry and texture.
Furthermore, we scale the radius of the high-quality pose to 2 and modify the field-of-view (FOV) angles to be consistent with the methodology employed by Zero-123-XL \cite{zero123}.
Regarding the generative model, we finetune Zero-123-XL \cite{zero123} on our collected Car360 dataset to improve the model generating more reliable car-specific prior.

In the geometry reconstruction stage, we set $\mathcal{L}_{rgb}=1000$, $\mathcal{L}_{mask}=100$, $\mathcal{L}_{depth}=0.05$, $\mathcal{L}_{normal}=1$, $\mathcal{L}_{mix}=0.1$.
In the texture refinement stage, to maximize the utilization of reference data, we set $\mathcal{L}_{rgb}=10000$, $\lambda_{3d}=0.1$, $\lambda_{booth}=0.1$, $\lambda_{lora}=0.01$. 
The pose optimization is only applied to the NeRF reconstruction stage with a learning rate of $10^{-5}$. Note that our pose optimization is integrated into NeRF training without explicit pose supervision.
Each stage is trained in 5000 epochs.
Completing all stages for a vehicle requires approximately 2 hours on an RTX 3090Ti GPU Card.

\noindent\textbf{Camera Processing.}   
Our method leverages Score Distillation Sampling and thus it requires the object centrally located within the conditioning image. Therefore, we re-center the object in an image based on its bounding box.

\subsection{Qualitative and Quantitative Results}
 \noindent\textbf{Datasets.} Our target is to reconstruct complete and fine 3D cars from the real-world scene. Therefore, we select 100 instances from our collected Car360 dataset for evaluation.  To imitate the moving forward scene, we randomly select 3 sparse views and 1 view from one side of each vehicle as training and testing views, respectively. To further demonstrate our method,  we also extract 100 vehicles from the Nuscenes \cite{nuscenes} dataset for further evaluation. The number of captured images for each vehicle ranges from 2 to 5. This dataset is documented in a moving-forward scenario, matching the realistic scene.

\noindent\textbf{Evaluation Settings and Metrics.} We evaluate the reconstruction quality by using image-based and lidar-based metrics. We use the common image-based metrics for rendered views, such as PSNR, SSIM, LPIPS, and FID. 
In image-based metrics, we denote the ``Standard'' as the evaluation in testing views. We also evaluate the mirror testing views denoted as ``Mirror'' in Table \ref{t1} to compare which method can reconstruct more complete 3D cars. 
For lidar-based metrics, we adopt per-ray $L_2$ error, Hit rate, Chamfer distance, and Hausdorff to measure the completeness of reconstructed 3D cars. Note that lidars in Nuscenes only scan one side of cars since it is in the moving forward scenes, so we would not evaluate this incomplete point cloud in Nuscenes.
On the contrary, our Car360 contains a complete point cloud for each vehicle, which would be used to evaluate the lidar-based metrics. 
In addition to these metrics, we also adopt map@0.5 as a detection metric to evaluate if existing detectors can detect reconstructed cars. Higher map@0.5 means the reconstructed results are more photorealistic.

\noindent\textbf{Car360.}  
As indicated in Table \ref{t1}, we adopt the recent 3D reconstruction methods, such as TensoRF \cite{TensoRF}, Instant-NGP \cite{instant-ngp}, and 3DGS \cite{kerbl2023gaussiansplatting} to compare with our proposed DreamCar. Notably, 3DGS \cite{kerbl2023gaussiansplatting}, despite being the recent state-of-the-art method, shows extremely bad quantitative metrics than ours.
When testing on the mirror views, their results are worse.
As shown in Table \ref{t2}, our proposed method obtains the best results against other methods in lidar-based metrics.
These methods struggle to fully reconstruct 3D objects when the test view significantly deviates from the reference.
We also compare our method with the recent 3D generation methods, such as Zero-123 \cite{zero123}, DreamGaussian \cite{tang2023dreamgaussian}, and DreamCraft3D \cite{sun2023dreamcraft3d}. For a fair comparison, we modify these models to accept multi-view inputs. Specifically, Dreamfusion \cite{dreamfusion} is utilized for Zero-123 in the 3D reconstruction process.
We also use existing detectors YOLOv8-l \cite{yolov8} to evaluate which method can reconstruct more photorealistic cars to be detected.
Specifically, we copy and paste the rendered images into the backgrounds of Nuscenes to form testing sets. Then, we use YOLOv8-l to detect these testing sets and only evaluate on the rendered objects.
The above comparisons indicate our DreamCar can produce better results than previous methods.

\noindent\textbf{Nuscenes.}   Table \ref{t3} depicts the quantitative results, where ``Standard'' and ``Mirror'' rows all show our method is superior to other 3D reconstruction methods. 
Previous 3D reconstruction methods fail to reconstruct cars given such a few images.
Our results in Nuscenes are worse than ours in Car360 since the images in Nuscenes are low-resolution, blurry, and noisy, which indicates this dataset is more challenging. 
As shown in  Figure \ref{vis_nus}, we show the visual comparison of 3D reconstruction with the recent 3D generation methods. Note that these methods are all modified to accept multi-view inputs.
3DGS can not reconstruct a complete 3D car given a few supervision images.
Zero-123 struggles to reconstruct a complete 3D geometry and detailed texture. The geometry reconstructed by DreamGussian is like an ellipsoid. Moreover, its texture is very blurry and over-smoothing.  DreamCraft3D, although a recent state-of-the-art 3D generation model, sometimes misses details such as vehicle wheels and produces textures that are overly saturated and lack photorealism.
Since our proposed method exploits mirror symmetry and has more powerful generalization in real-world cars, our DreamCar can reconstruct real-world cars from the moving forward scene. These comparisons demonstrate the applicability of our method in realistic scenes for the production of large-scale 3D car assets.

\subsection{Ablation Study}

In Figure \ref{ablation}, we show the ablation study through ablating our proposed mirror and pose optimization, respectively. When our full model does not use mirror symmetry, the generated geometry is incomplete and the texture looks different with the reference. Without pose optimization, the generated texture displays misalignments in the vehicle wheels. To highlight the benefits of our Car360 dataset, we conduct a novel view synthesis comparison in Figure \ref{vis_nvs}, with other state-of-the-art 3D-aware diffusion models, such as Zero-123-XL \cite{zero123}, SyncDreamer \cite{liu2023syncdreamer}, and Stable-Zero123 \cite{stable-zero123}.  As we can see, Zero-123-XL and SyncDreamer generate unrealistic results and seem to pattern collapse.  Stable-Zero123 also generates unreasonable features, as the vehicle window is not matched with the real one. Moreover, it seems the generating pose has a few deviations. The generative model trained on our Car360 dataset obtains powerful generalization to real cars and accomplishes better visual results than others. These comparisons demonstrate the necessity of our collected Car360 dataset.


\subsection{Car Simulation.}
As illustrated in Figure \ref{vis_simu}, we present the visualization of the car simulation. We first use our method to reconstruct a bulk of 3D car models. Then, the reconstructed 3D car models are employed for simulation within Nuscenes \cite{nuscenes} scenes. Utilizing these 3D cars allows for the simulation of large-scale, highly hazardous scenarios in a cost-effective way.

\section{Conclusion}
\vspace{-2mm}
In this work, we propose a groundbreaking 3D car reconstruction method, named DreamCar. 
The success of DreamCar can be attributed to its innovative application of mirror symmetry, car-specific generative prior, and pose optimization techniques. Specifically, the usage of the collected Car360 datasets improves the generalization of our method to real-world cars.  We demonstrate our DreamCar can effectively reconstruct large-scale exquisite 3D car assets and outperform other state-of-the-art methods.



\bibliographystyle{ieeetr}
\bibliography{egbib}

\begin{thebibliography}{10}

\bibitem{tan2021scenegen}
S.~Tan, K.~Wong, S.~Wang, S.~Manivasagam, M.~Ren, and R.~Urtasun, ``Scenegen: Learning to generate realistic traffic scenes,'' in {\em Proceedings of the IEEE/CVF Conference on Computer Vision and Pattern Recognition}, pp.~892--901, 2021.

\bibitem{nuscenes}
H.~Caesar, V.~Bankiti, A.~H. Lang, S.~Vora, V.~E. Liong, Q.~Xu, A.~Krishnan, Y.~Pan, G.~Baldan, and O.~Beijbom, ``nuscenes: A multimodal dataset for autonomous driving,'' in {\em Proceedings of the IEEE/CVF conference on computer vision and pattern recognition}, pp.~11621--11631, 2020.

\bibitem{waymo}
P.~Sun, H.~Kretzschmar, X.~Dotiwalla, A.~Chouard, V.~Patnaik, P.~Tsui, J.~Guo, Y.~Zhou, Y.~Chai, B.~Caine, {\em et~al.}, ``Scalability in perception for autonomous driving: Waymo open dataset,'' in {\em Proceedings of the IEEE/CVF conference on computer vision and pattern recognition}, pp.~2446--2454, 2020.

\bibitem{kitti}
A.~Geiger, P.~Lenz, C.~Stiller, and R.~Urtasun, ``Vision meets robotics: The kitti dataset,'' {\em The International Journal of Robotics Research}, vol.~32, no.~11, pp.~1231--1237, 2013.

\bibitem{zero123}
R.~Liu, R.~Wu, B.~Van~Hoorick, P.~Tokmakov, S.~Zakharov, and C.~Vondrick, ``Zero-1-to-3: Zero-shot one image to 3d object,'' in {\em Proceedings of the IEEE/CVF International Conference on Computer Vision}, pp.~9298--9309, 2023.

\bibitem{liu2023syncdreamer}
Y.~Liu, C.~Lin, Z.~Zeng, X.~Long, L.~Liu, T.~Komura, and W.~Wang, ``Syncdreamer: Generating multiview-consistent images from a single-view image,'' {\em arXiv preprint arXiv:2309.03453}, 2023.

\bibitem{stable-zero123}
S.~AI, ``{Stable Zero123}: Quality 3d object generation from single images.'' \url{https://stability.ai/news/stable-zero123-3d-generation}, 2023.

\bibitem{zhang2020nerf++}
K.~Zhang, G.~Riegler, N.~Snavely, and V.~Koltun, ``Nerf++: Analyzing and improving neural radiance fields,'' {\em arXiv preprint arXiv:2010.07492}, 2020.

\bibitem{tancik2022block}
M.~Tancik, V.~Casser, X.~Yan, S.~Pradhan, B.~Mildenhall, P.~P. Srinivasan, J.~T. Barron, and H.~Kretzschmar, ``Block-nerf: Scalable large scene neural view synthesis,'' in {\em Proceedings of the IEEE/CVF Conference on Computer Vision and Pattern Recognition}, pp.~8248--8258, 2022.

\bibitem{xian2021space}
W.~Xian, J.-B. Huang, J.~Kopf, and C.~Kim, ``Space-time neural irradiance fields for free-viewpoint video,'' in {\em Proceedings of the IEEE/CVF Conference on Computer Vision and Pattern Recognition}, pp.~9421--9431, 2021.

\bibitem{boss2021nerd}
M.~Boss, R.~Braun, V.~Jampani, J.~T. Barron, C.~Liu, and H.~Lensch, ``Nerd: Neural reflectance decomposition from image collections,'' in {\em Proceedings of the IEEE/CVF International Conference on Computer Vision}, pp.~12684--12694, 2021.

\bibitem{wei2021nerfingmvs}
Y.~Wei, S.~Liu, Y.~Rao, W.~Zhao, J.~Lu, and J.~Zhou, ``Nerfingmvs: Guided optimization of neural radiance fields for indoor multi-view stereo,'' in {\em Proceedings of the IEEE/CVF International Conference on Computer Vision}, pp.~5610--5619, 2021.

\bibitem{reiser2023merf}
C.~Reiser, R.~Szeliski, D.~Verbin, P.~Srinivasan, B.~Mildenhall, A.~Geiger, J.~Barron, and P.~Hedman, ``Merf: Memory-efficient radiance fields for real-time view synthesis in unbounded scenes,'' {\em ACM Transactions on Graphics (TOG)}, vol.~42, no.~4, pp.~1--12, 2023.

\bibitem{yuan2022nerf}
Y.-J. Yuan, Y.-T. Sun, Y.-K. Lai, Y.~Ma, R.~Jia, and L.~Gao, ``Nerf-editing: geometry editing of neural radiance fields,'' in {\em Proceedings of the IEEE/CVF Conference on Computer Vision and Pattern Recognition}, pp.~18353--18364, 2022.

\bibitem{garbin2021fastnerf}
S.~J. Garbin, M.~Kowalski, M.~Johnson, J.~Shotton, and J.~Valentin, ``Fastnerf: High-fidelity neural rendering at 200fps,'' in {\em Proceedings of the IEEE/CVF International Conference on Computer Vision}, pp.~14346--14355, 2021.

\bibitem{park2021nerfies}
K.~Park, U.~Sinha, J.~T. Barron, S.~Bouaziz, D.~B. Goldman, S.~M. Seitz, and R.~Martin-Brualla, ``Nerfies: Deformable neural radiance fields,'' in {\em Proceedings of the IEEE/CVF International Conference on Computer Vision}, pp.~5865--5874, 2021.

\bibitem{chen2023mobilenerf}
Z.~Chen, T.~Funkhouser, P.~Hedman, and A.~Tagliasacchi, ``Mobilenerf: Exploiting the polygon rasterization pipeline for efficient neural field rendering on mobile architectures,'' in {\em Proceedings of the IEEE/CVF Conference on Computer Vision and Pattern Recognition}, pp.~16569--16578, 2023.

\bibitem{lin2021barf}
C.-H. Lin, W.-C. Ma, A.~Torralba, and S.~Lucey, ``Barf: Bundle-adjusting neural radiance fields,'' in {\em Proceedings of the IEEE/CVF International Conference on Computer Vision}, pp.~5741--5751, 2021.

\bibitem{tancik2023nerfstudio}
M.~Tancik, E.~Weber, E.~Ng, R.~Li, B.~Yi, T.~Wang, A.~Kristoffersen, J.~Austin, K.~Salahi, A.~Ahuja, {\em et~al.}, ``Nerfstudio: A modular framework for neural radiance field development,'' in {\em ACM SIGGRAPH 2023 Conference Proceedings}, pp.~1--12, 2023.

\bibitem{mildenhall2021nerf}
B.~Mildenhall, P.~P. Srinivasan, M.~Tancik, J.~T. Barron, R.~Ramamoorthi, and R.~Ng, ``Nerf: Representing scenes as neural radiance fields for view synthesis,'' {\em Communications of the ACM}, vol.~65, no.~1, pp.~99--106, 2021.

\bibitem{volume_rendering}
J.~T. Kajiya and B.~P. Von~Herzen, ``Ray tracing volume densities,'' {\em ACM SIGGRAPH computer graphics}, vol.~18, no.~3, pp.~165--174, 1984.

\bibitem{mip-nerf360}
J.~T. Barron, B.~Mildenhall, D.~Verbin, P.~P. Srinivasan, and P.~Hedman, ``Mip-nerf 360: Unbounded anti-aliased neural radiance fields,'' in {\em Proceedings of the IEEE/CVF Conference on Computer Vision and Pattern Recognition}, pp.~5470--5479, 2022.

\bibitem{mip-nerf}
J.~T. Barron, B.~Mildenhall, M.~Tancik, P.~Hedman, R.~Martin-Brualla, and P.~P. Srinivasan, ``Mip-nerf: A multiscale representation for anti-aliasing neural radiance fields,'' in {\em Proceedings of the IEEE/CVF International Conference on Computer Vision}, pp.~5855--5864, 2021.

\bibitem{TensoRF}
A.~Chen, Z.~Xu, A.~Geiger, J.~Yu, and H.~Su, ``Tensorf: Tensorial radiance fields,'' {\em arXiv preprint arXiv:2203.09517}, 2022.

\bibitem{instant-ngp}
T.~M\"uller, A.~Evans, C.~Schied, and A.~Keller, ``Instant neural graphics primitives with a multiresolution hash encoding,'' {\em arXiv:2201.05989}, Jan. 2022.

\bibitem{kerbl2023gaussiansplatting}
B.~Kerbl, G.~Kopanas, T.~Leimk{\"u}hler, and G.~Drettakis, ``3d gaussian splatting for real-time radiance field rendering,'' {\em ACM Transactions on Graphics (ToG)}, vol.~42, no.~4, pp.~1--14, 2023.

\bibitem{stablediffusion}
R.~Rombach, A.~Blattmann, D.~Lorenz, P.~Esser, and B.~Ommer, ``High-resolution image synthesis with latent diffusion models,'' in {\em Proceedings of the IEEE/CVF conference on computer vision and pattern recognition}, pp.~10684--10695, 2022.

\bibitem{dhariwal2021diffusion}
P.~Dhariwal and A.~Nichol, ``Diffusion models beat gans on image synthesis,'' {\em Advances in neural information processing systems}, vol.~34, pp.~8780--8794, 2021.

\bibitem{dreamfusion}
B.~Poole, A.~Jain, J.~T. Barron, and B.~Mildenhall, ``Dreamfusion: Text-to-3d using 2d diffusion,'' {\em arXiv preprint arXiv:2209.14988}, 2022.

\bibitem{sjc}
H.~Wang, X.~Du, J.~Li, R.~A. Yeh, and G.~Shakhnarovich, ``Score jacobian chaining: Lifting pretrained 2d diffusion models for 3d generation,'' in {\em Proceedings of the IEEE/CVF Conference on Computer Vision and Pattern Recognition}, pp.~12619--12629, 2023.

\bibitem{chen2023text}
Z.~Chen, F.~Wang, and H.~Liu, ``Text-to-3d using gaussian splatting,'' {\em arXiv preprint arXiv:2309.16585}, 2023.

\bibitem{seo2023ditto}
H.~Seo, H.~Kim, G.~Kim, and S.~Y. Chun, ``Ditto-nerf: Diffusion-based iterative text to omni-directional 3d model,'' {\em arXiv preprint arXiv:2304.02827}, 2023.

\bibitem{yu2023points}
C.~Yu, Q.~Zhou, J.~Li, Z.~Zhang, Z.~Wang, and F.~Wang, ``Points-to-3d: Bridging the gap between sparse points and shape-controllable text-to-3d generation,'' {\em arXiv preprint arXiv:2307.13908}, 2023.

\bibitem{seo2023let}
J.~Seo, W.~Jang, M.-S. Kwak, J.~Ko, H.~Kim, J.~Kim, J.-H. Kim, J.~Lee, and S.~Kim, ``Let 2d diffusion model know 3d-consistency for robust text-to-3d generation,'' {\em arXiv preprint arXiv:2303.07937}, 2023.

\bibitem{tsalicoglou2023textmesh}
C.~Tsalicoglou, F.~Manhardt, A.~Tonioni, M.~Niemeyer, and F.~Tombari, ``Textmesh: Generation of realistic 3d meshes from text prompts,'' {\em arXiv preprint arXiv:2304.12439}, 2023.

\bibitem{armandpour2023re}
M.~Armandpour, H.~Zheng, A.~Sadeghian, A.~Sadeghian, and M.~Zhou, ``Re-imagine the negative prompt algorithm: Transform 2d diffusion into 3d, alleviate janus problem and beyond,'' {\em arXiv preprint arXiv:2304.04968}, 2023.

\bibitem{chen2023it3d}
Y.~Chen, C.~Zhang, X.~Yang, Z.~Cai, G.~Yu, L.~Yang, and G.~Lin, ``It3d: Improved text-to-3d generation with explicit view synthesis,'' {\em arXiv preprint arXiv:2308.11473}, 2023.

\bibitem{xu2022neurallift}
D.~Xu, Y.~Jiang, P.~Wang, Z.~Fan, Y.~Wang, and Z.~Wang, ``Neurallift-360: Lifting an in-the-wild 2d photo to a 3d object with 360 views,'' {\em arXiv e-prints}, pp.~arXiv--2211, 2022.

\bibitem{cheng2023progressive3d}
X.~Cheng, T.~Yang, J.~Wang, Y.~Li, L.~Zhang, J.~Zhang, and L.~Yuan, ``Progressive3d: Progressively local editing for text-to-3d content creation with complex semantic prompts,'' {\em arXiv preprint arXiv:2310.11784}, 2023.

\bibitem{qian2023magic123}
G.~Qian, J.~Mai, A.~Hamdi, J.~Ren, A.~Siarohin, B.~Li, H.-Y. Lee, I.~Skorokhodov, P.~Wonka, S.~Tulyakov, {\em et~al.}, ``Magic123: One image to high-quality 3d object generation using both 2d and 3d diffusion priors,'' {\em arXiv preprint arXiv:2306.17843}, 2023.

\bibitem{yu2023hifi}
W.~Yu, L.~Yuan, Y.-P. Cao, X.~Gao, X.~Li, L.~Quan, Y.~Shan, and Y.~Tian, ``Hifi-123: Towards high-fidelity one image to 3d content generation,'' {\em arXiv preprint arXiv:2310.06744}, 2023.

\bibitem{shen2023anything}
Q.~Shen, X.~Yang, and X.~Wang, ``Anything-3d: Towards single-view anything reconstruction in the wild,'' {\em arXiv preprint arXiv:2304.10261}, 2023.

\bibitem{melas2023realfusion}
L.~Melas-Kyriazi, I.~Laina, C.~Rupprecht, and A.~Vedaldi, ``Realfusion: 360deg reconstruction of any object from a single image,'' in {\em CVPR}, 2023.

\bibitem{tang2023make}
J.~Tang, T.~Wang, B.~Zhang, T.~Zhang, R.~Yi, L.~Ma, and D.~Chen, ``Make-it-3d: High-fidelity 3d creation from a single image with diffusion prior,'' in {\em ICCV}, 2023.

\bibitem{ling2023align}
H.~Ling, S.~W. Kim, A.~Torralba, S.~Fidler, and K.~Kreis, ``Align your gaussians: Text-to-4d with dynamic 3d gaussians and composed diffusion models,'' {\em arXiv preprint arXiv:2312.13763}, 2023.

\bibitem{ma2023x}
Y.~Ma, Y.~Fan, J.~Ji, H.~Wang, X.~Sun, G.~Jiang, A.~Shu, and R.~Ji, ``X-dreamer: Creating high-quality 3d content by bridging the domain gap between text-to-2d and text-to-3d generation,'' {\em arXiv preprint arXiv:2312.00085}, 2023.

\bibitem{shi2023mvdream}
Y.~Shi, P.~Wang, J.~Ye, M.~Long, K.~Li, and X.~Yang, ``Mvdream: Multi-view diffusion for 3d generation,'' {\em arXiv preprint arXiv:2308.16512}, 2023.

\bibitem{li2023sweetdreamer}
W.~Li, R.~Chen, X.~Chen, and P.~Tan, ``Sweetdreamer: Aligning geometric priors in 2d diffusion for consistent text-to-3d,'' {\em arXiv preprint arXiv:2310.02596}, 2023.

\bibitem{huang2023dreamcontrol}
T.~Huang, Y.~Zeng, Z.~Zhang, W.~Xu, H.~Xu, S.~Xu, R.~W. Lau, and W.~Zuo, ``Dreamcontrol: Control-based text-to-3d generation with 3d self-prior,'' {\em arXiv preprint arXiv:2312.06439}, 2023.

\bibitem{long2023wonder3d}
X.~Long, Y.-C. Guo, C.~Lin, Y.~Liu, Z.~Dou, L.~Liu, Y.~Ma, S.-H. Zhang, M.~Habermann, C.~Theobalt, {\em et~al.}, ``Wonder3d: Single image to 3d using cross-domain diffusion,'' {\em arXiv preprint arXiv:2310.15008}, 2023.

\bibitem{wang2023prolificdreamer}
Z.~Wang, C.~Lu, Y.~Wang, F.~Bao, C.~Li, H.~Su, and J.~Zhu, ``Prolificdreamer: High-fidelity and diverse text-to-3d generation with variational score distillation,'' {\em arXiv preprint arXiv:2305.16213}, 2023.

\bibitem{zhang2023repaint123}
J.~Zhang, Z.~Tang, Y.~Pang, X.~Cheng, P.~Jin, Y.~Wei, W.~Yu, M.~Ning, and L.~Yuan, ``Repaint123: Fast and high-quality one image to 3d generation with progressive controllable 2d repainting,'' {\em arXiv preprint arXiv:2312.13271}, 2023.

\bibitem{szymanowicz2023viewset}
S.~Szymanowicz, C.~Rupprecht, and A.~Vedaldi, ``Viewset diffusion:(0-) image-conditioned 3d generative models from 2d data,'' {\em arXiv preprint arXiv:2306.07881}, 2023.

\bibitem{liu2023one2345++}
M.~Liu, R.~Shi, L.~Chen, Z.~Zhang, C.~Xu, X.~Wei, H.~Chen, C.~Zeng, J.~Gu, and H.~Su, ``One-2-3-45++: Fast single image to 3d objects with consistent multi-view generation and 3d diffusion,'' {\em arXiv preprint arXiv:2311.07885}, 2023.

\bibitem{shi2023zero123plus}
R.~Shi, H.~Chen, Z.~Zhang, M.~Liu, C.~Xu, X.~Wei, L.~Chen, C.~Zeng, and H.~Su, ``Zero123++: a single image to consistent multi-view diffusion base model,'' 2023.

\bibitem{raj2023dreambooth3d}
A.~Raj, S.~Kaza, B.~Poole, M.~Niemeyer, N.~Ruiz, B.~Mildenhall, S.~Zada, K.~Aberman, M.~Rubinstein, J.~Barron, {\em et~al.}, ``Dreambooth3d: Subject-driven text-to-3d generation,'' {\em arXiv preprint arXiv:2303.13508}, 2023.

\bibitem{chen2023fantasia3d}
R.~Chen, Y.~Chen, N.~Jiao, and K.~Jia, ``Fantasia3d: Disentangling geometry and appearance for high-quality text-to-3d content creation,'' {\em arXiv preprint arXiv:2303.13873}, 2023.

\bibitem{liu2024one}
M.~Liu, C.~Xu, H.~Jin, L.~Chen, M.~Varma~T, Z.~Xu, and H.~Su, ``One-2-3-45: Any single image to 3d mesh in 45 seconds without per-shape optimization,'' {\em Advances in Neural Information Processing Systems}, vol.~36, 2024.

\bibitem{lin2023magic3d}
C.-H. Lin, J.~Gao, L.~Tang, T.~Takikawa, X.~Zeng, X.~Huang, K.~Kreis, S.~Fidler, M.-Y. Liu, and T.-Y. Lin, ``Magic3d: High-resolution text-to-3d content creation,'' in {\em CVPR}, 2023.

\bibitem{liu2023unidream}
Z.~Liu, Y.~Li, Y.~Lin, X.~Yu, S.~Peng, Y.-P. Cao, X.~Qi, X.~Huang, D.~Liang, and W.~Ouyang, ``Unidream: Unifying diffusion priors for relightable text-to-3d generation,'' {\em arXiv preprint arXiv:2312.08754}, 2023.

\bibitem{zhao2023efficientdreamer}
M.~Zhao, C.~Zhao, X.~Liang, L.~Li, Z.~Zhao, Z.~Hu, C.~Fan, and X.~Yu, ``Efficientdreamer: High-fidelity and robust 3d creation via orthogonal-view diffusion prior,'' {\em arXiv preprint arXiv:2308.13223}, 2023.

\bibitem{zhu2023hifa}
J.~Zhu and P.~Zhuang, ``Hifa: High-fidelity text-to-3d with advanced diffusion guidance,'' {\em arXiv preprint arXiv:2305.18766}, 2023.

\bibitem{huang2023dreamtime}
Y.~Huang, J.~Wang, Y.~Shi, X.~Qi, Z.-J. Zha, and L.~Zhang, ``Dreamtime: An improved optimization strategy for text-to-3d content creation,'' {\em arXiv preprint arXiv:2306.12422}, 2023.

\bibitem{wu2023hd}
J.~Wu, X.~Gao, X.~Liu, Z.~Shen, C.~Zhao, H.~Feng, J.~Liu, and E.~Ding, ``Hd-fusion: Detailed text-to-3d generation leveraging multiple noise estimation,'' {\em arXiv preprint arXiv:2307.16183}, 2023.

\bibitem{jiang2023efficient}
Y.~Jiang, H.~Tang, J.-H.~R. Chang, L.~Song, Z.~Wang, and L.~Cao, ``Efficient-3dim: Learning a generalizable single-image novel-view synthesizer in one day,'' {\em arXiv preprint arXiv:2310.03015}, 2023.

\bibitem{chen2023et3d}
Y.~Chen, Z.~Li, and P.~Liu, ``Et3d: Efficient text-to-3d generation via multi-view distillation,'' {\em arXiv preprint arXiv:2311.15561}, 2023.

\bibitem{yi2023gaussiandreamer}
T.~Yi, J.~Fang, G.~Wu, L.~Xie, X.~Zhang, W.~Liu, Q.~Tian, and X.~Wang, ``Gaussiandreamer: Fast generation from text to 3d gaussian splatting with point cloud priors,'' {\em arXiv preprint arXiv:2310.08529}, 2023.

\bibitem{liu2023humangaussian}
X.~Liu, X.~Zhan, J.~Tang, Y.~Shan, G.~Zeng, D.~Lin, X.~Liu, and Z.~Liu, ``Humangaussian: Text-driven 3d human generation with gaussian splatting,'' {\em arXiv preprint arXiv:2311.17061}, 2023.

\bibitem{chung2023luciddreamer}
J.~Chung, S.~Lee, H.~Nam, J.~Lee, and K.~M. Lee, ``Luciddreamer: Domain-free generation of 3d gaussian splatting scenes,'' {\em arXiv preprint arXiv:2311.13384}, 2023.

\bibitem{xu2024agg}
D.~Xu, Y.~Yuan, M.~Mardani, S.~Liu, J.~Song, Z.~Wang, and A.~Vahdat, ``Agg: Amortized generative 3d gaussians for single image to 3d,'' {\em arXiv preprint arXiv:2401.04099}, 2024.

\bibitem{szymanowicz2023splatter}
S.~Szymanowicz, C.~Rupprecht, and A.~Vedaldi, ``Splatter image: Ultra-fast single-view 3d reconstruction,'' {\em arXiv preprint arXiv:2312.13150}, 2023.

\bibitem{tang2023dreamgaussian}
J.~Tang, J.~Ren, H.~Zhou, Z.~Liu, and G.~Zeng, ``Dreamgaussian: Generative gaussian splatting for efficient 3d content creation,'' {\em arXiv preprint arXiv:2309.16653}, 2023.

\bibitem{sun2023dreamcraft3d}
J.~Sun, B.~Zhang, R.~Shao, L.~Wang, W.~Liu, Z.~Xie, and Y.~Liu, ``Dreamcraft3d: Hierarchical 3d generation with bootstrapped diffusion prior,'' {\em arXiv preprint arXiv:2310.16818}, 2023.

\bibitem{deitke2023objaverse}
M.~Deitke, D.~Schwenk, J.~Salvador, L.~Weihs, O.~Michel, E.~VanderBilt, L.~Schmidt, K.~Ehsani, A.~Kembhavi, and A.~Farhadi, ``Objaverse: A universe of annotated 3d objects,'' in {\em Proceedings of the IEEE/CVF Conference on Computer Vision and Pattern Recognition}, pp.~13142--13153, 2023.

\bibitem{deitke2024objaverse}
M.~Deitke, R.~Liu, M.~Wallingford, H.~Ngo, O.~Michel, A.~Kusupati, A.~Fan, C.~Laforte, V.~Voleti, S.~Y. Gadre, {\em et~al.}, ``Objaverse-xl: A universe of 10m+ 3d objects,'' {\em Advances in Neural Information Processing Systems}, vol.~36, 2024.

\bibitem{chang2015shapenet}
A.~X. Chang, T.~Funkhouser, L.~Guibas, P.~Hanrahan, Q.~Huang, Z.~Li, S.~Savarese, M.~Savva, S.~Song, H.~Su, {\em et~al.}, ``Shapenet: An information-rich 3d model repository,'' {\em arXiv preprint arXiv:1512.03012}, 2015.

\bibitem{poole2022dreamfusion}
B.~Poole, A.~Jain, J.~T. Barron, and B.~Mildenhall, ``Dreamfusion: Text-to-3d using 2d diffusion,'' {\em arXiv preprint arXiv:2209.14988}, 2022.

\bibitem{barron2021mip}
J.~T. Barron, B.~Mildenhall, M.~Tancik, P.~Hedman, R.~Martin-Brualla, and P.~P. Srinivasan, ``Mip-nerf: A multiscale representation for anti-aliasing neural radiance fields,'' in {\em Proceedings of the IEEE/CVF International Conference on Computer Vision}, pp.~5855--5864, 2021.

\bibitem{sfm2016}
J.~L. Schonberger and J.-M. Frahm, ``Structure-from-motion revisited,'' in {\em 2016 IEEE Conference on Computer Vision and Pattern Recognition (CVPR)}, Jun 2016.

\bibitem{mishra2013lie}
A.~Mishra, S.~P. Awate, A.~Banerjee, N.~El-Zehiry, S.~Kurtek, S.~J. McKenna, and A.~Tannenbaum, ``Lie groups in computer vision and image processing: A survey,'' {\em Journal of Mathematical Imaging and Vision}, vol.~47, no.~3, pp.~209--252, 2013.

\bibitem{sam}
A.~Kirillov, E.~Mintun, N.~Ravi, H.~Mao, C.~Rolland, L.~Gustafson, T.~Xiao, S.~Whitehead, A.~C. Berg, W.-Y. Lo, {\em et~al.}, ``Segment anything,'' {\em arXiv preprint arXiv:2304.02643}, 2023.

\bibitem{wang2021neus}
P.~Wang, L.~Liu, Y.~Liu, C.~Theobalt, T.~Komura, and W.~Wang, ``Neus: Learning neural implicit surfaces by volume rendering for multi-view reconstruction,'' {\em arXiv preprint arXiv:2106.10689}, 2021.

\bibitem{shen2021deep}
T.~Shen, J.~Gao, K.~Yin, M.-Y. Liu, and S.~Fidler, ``Deep marching tetrahedra: a hybrid representation for high-resolution 3d shape synthesis,'' {\em Advances in Neural Information Processing Systems}, vol.~34, pp.~6087--6101, 2021.

\bibitem{deng2023nerdi}
C.~Deng, C.~Jiang, C.~R. Qi, X.~Yan, Y.~Zhou, L.~Guibas, D.~Anguelov, {\em et~al.}, ``Nerdi: Single-view nerf synthesis with language-guided diffusion as general image priors,'' in {\em Proceedings of the IEEE/CVF Conference on Computer Vision and Pattern Recognition}, pp.~20637--20647, 2023.

\bibitem{eftekhar2021omnidata}
A.~Eftekhar, A.~Sax, J.~Malik, and A.~Zamir, ``Omnidata: A scalable pipeline for making multi-task mid-level vision datasets from 3d scans,'' in {\em Proceedings of the IEEE/CVF International Conference on Computer Vision}, pp.~10786--10796, 2021.

\bibitem{deep-floyd}
A.~Shonenkov, M.~Konstantinov, D.~Bakshandaeva, C.~Schuhmann, K.~Ivanova, and N.~Klokova, ``{DeepFloyd IF}: A modular cascaded diffusion model.'' \url{https://github.com/deep-floyd/IF/tree/develop}, 2023.

\bibitem{ruiz2023dreambooth}
N.~Ruiz, Y.~Li, V.~Jampani, Y.~Pritch, M.~Rubinstein, and K.~Aberman, ``Dreambooth: Fine tuning text-to-image diffusion models for subject-driven generation,'' in {\em Proceedings of the IEEE/CVF Conference on Computer Vision and Pattern Recognition}, pp.~22500--22510, 2023.

\bibitem{hu2021lora}
E.~J. Hu, Y.~Shen, P.~Wallis, Z.~Allen-Zhu, Y.~Li, S.~Wang, L.~Wang, and W.~Chen, ``Lora: Low-rank adaptation of large language models,'' {\em arXiv preprint arXiv:2106.09685}, 2021.

\bibitem{barron2023zipnerf}
J.~T. Barron, B.~Mildenhall, D.~Verbin, P.~P. Srinivasan, and P.~Hedman, ``Zip-nerf: Anti-aliased grid-based neural radiance fields,'' {\em ICCV}, 2023.

\bibitem{yolov8}
Ultralytics, ``{YOLOv8}: A cutting-edge and state-of-the-art (sota) model that builds upon the success of previous yolo versions.'' \url{https://github.com/ultralytics/ultralytics?tab=readme-ov-file }, 2023.

\end{thebibliography}


\newpage

 


\begin{IEEEbiography}[{\includegraphics[width=1in,height=1.25in,clip,keepaspectratio]{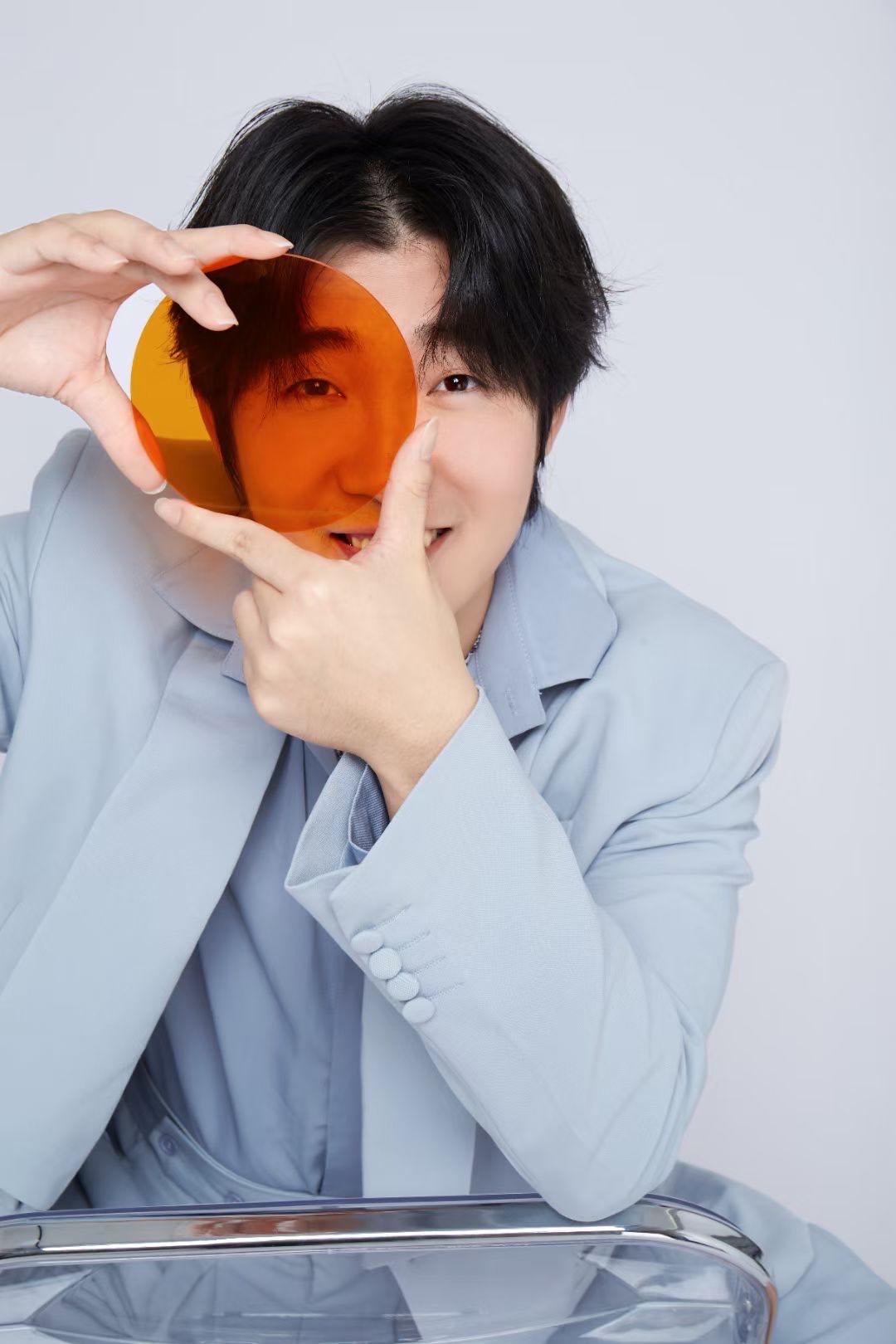}}]{Xiaobiao Du}
is currently pursuing a Ph.D degree at the University of Technology Sydney, Australia. His research interests include 3D vision and generation. He is particularly interested in improving few-shot 3D reconstruction with generative prior.
\end{IEEEbiography}

\begin{IEEEbiography}[{\includegraphics[width=1in,height=1.25in,clip,keepaspectratio]{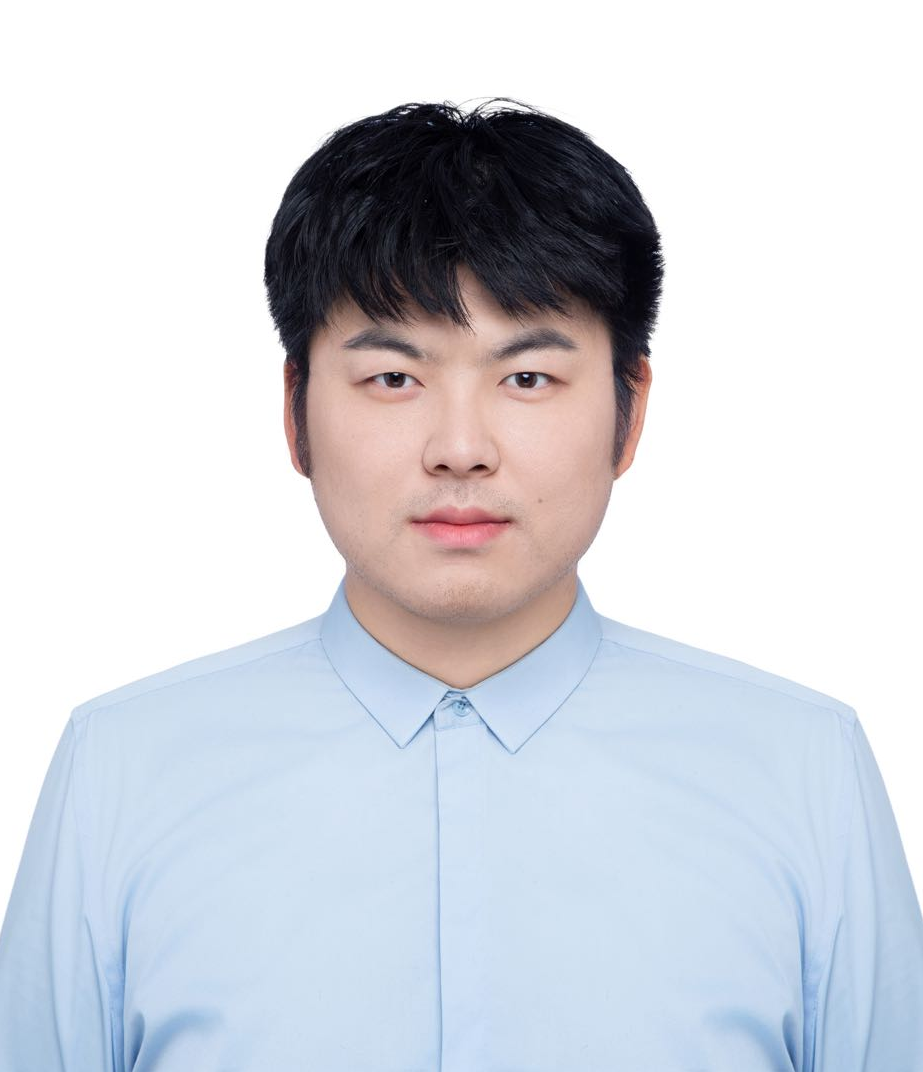}}]{Haiyang Sun}
received the BS degree in information and communication engineering from Tsinghua University, Beijing, China, in 2016.  He is currently an algorithm expert of LiAuto's Autonomous Driving Team. His research areas include perception algorithms, AIGC, and the neural field.
\end{IEEEbiography}

\begin{IEEEbiography}[{\includegraphics[width=1in,height=1.25in,clip,keepaspectratio]{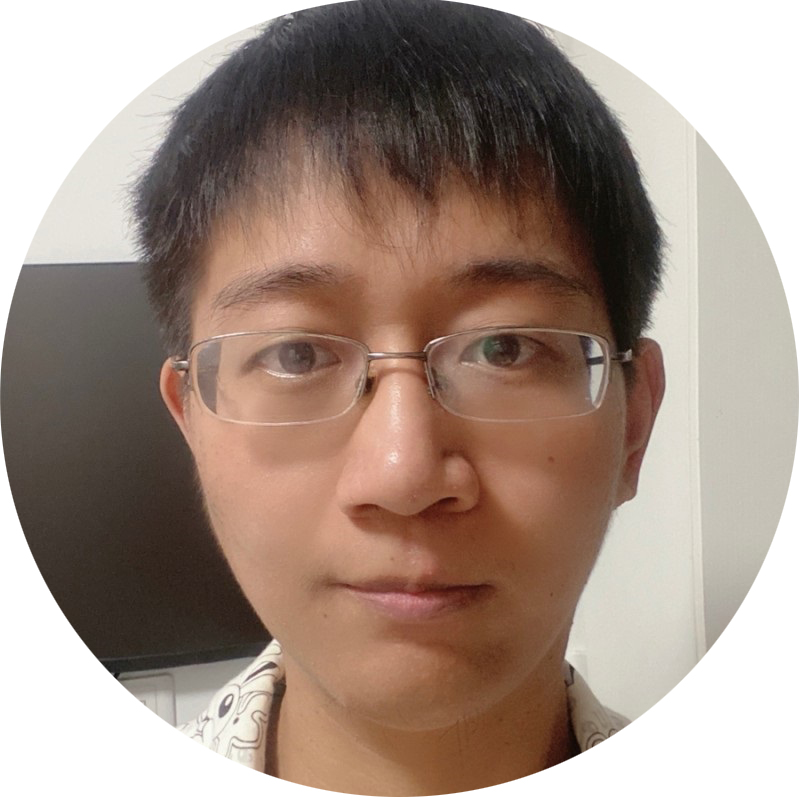}}]{Ming Lu}
received the Ph.D. degree in information and communication engineering from Tsinghua University, Beijing, China, in 2019. He is currently a Staff Researcher with Intel Labs China, Beijing. His research interests include 3D vision and computer graphics. He is particularly interested in improving the workloads at crucial visual synthesis systems, such as AI + Chips (ISP/Codec/GPU), AIGC, neural field, and large AI models.
\end{IEEEbiography}

\begin{IEEEbiography}[{\includegraphics[width=1in,height=1.25in,clip,keepaspectratio]{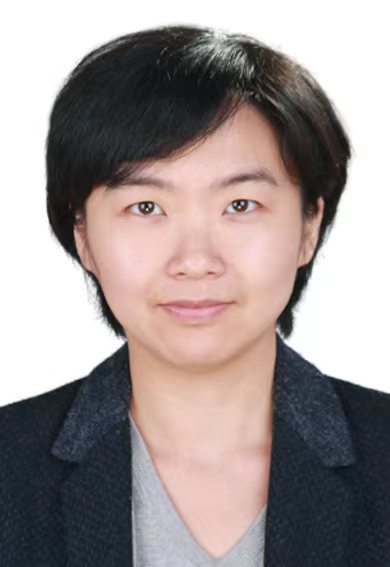}}]{Tianqing Zhu}
 received her B.Eng. degree and her M.Eng. degree from Wuhan University, China, in 2000 and 2004, respectively. She also holds a PhD in computer science from Deakin University, Australia (2014). She is currently a professor at city university of Macau. Prior to that, she was a lecturer with the School of Information Technology, Deakin University, associate professor at Sydney University of Technology. Her research interests include privacy preserving, AI security and privacy, and network security.
\end{IEEEbiography}

\begin{IEEEbiography}[{\includegraphics[width=1in,height=1.25in,clip,keepaspectratio]{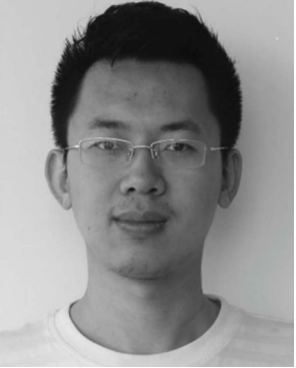}}]{Xin Yu}
received the BS degree in electronic engineering from the University of Electronic Science
and Technology of China, Chengdu, China, in 2009,
the PhD degree from the Department of Electronic
Engineering, Tsinghua University, Beijing, China,
in 2015, and the PhD degree from the College of
Engineering and Computer Science, Australian National University, Canberra, Australia, in 2019. He
is currently a senior lecturer with the University of
Queensland. His research interests include computer
vision and image processing.
\end{IEEEbiography}

\vfill

\end{document}